\documentclass[]{fairmeta}

\usepackage{amsmath}
\usepackage{amssymb}
\usepackage{bbm}
\usepackage{verbatim}
\usepackage{wrapfig}
\usepackage{subcaption}

\newtcolorbox[auto counter, crefname={Example}{Examples}, Crefname={Example}{Examples}]{examplebox}[2][]{
  colback=gray!5,
  colframe=gray!40,
  fonttitle=\sffamily\bfseries,
  title={Example~\thetcbcounter: #2},
  label={#1},
  breakable
}

\newtcolorbox{promptbox}[1][]{
  colback=gray!5,
  colframe=gray!40,
  fonttitle=\sffamily\bfseries,
  title={#1},
  breakable
}

\newtcolorbox{untunedbox}{
  breakable=false, enhanced,
  colback=red!5, colbacktitle=red!12, colframe=red!30!gray!70, coltitle=black,
  fonttitle=\sffamily\bfseries\small, boxrule=0.4pt, arc=2pt,
  left=4pt, right=4pt, top=2pt, bottom=2pt, before skip=4pt, after skip=6pt,
  title={Untuned 7B CG}
}
\newtcolorbox{trainedbox}{
  breakable=false, enhanced,
  colback=green!7, colbacktitle=green!16, colframe=green!45!black!40, coltitle=black,
  fonttitle=\sffamily\bfseries\small, boxrule=0.4pt, arc=2pt,
  left=4pt, right=4pt, top=2pt, bottom=2pt, before skip=4pt, after skip=6pt,
  title={Trained max-of-mean}
}

\title{Beyond Repeated Sampling: Learning Search Policies for LLM Reasoning
}

\author[1,2]{Ismail Labiad}
\author[2]{Matthieu Kowalski}
\author[2]{Marc Schoenauer}
\author[1]{Rémi Munos}
\author[1,3]{Julia Kempe}

\affiliation[1]{Meta FAIR}
\affiliation[2]{Université Paris-Saclay, LISN, Inria, CNRS}
\affiliation[3]{NYU Courant Institute and CDS}

\abstract{
Large language models increasingly tackle hard reasoning problems by spending more test-time compute, yet the dominant strategy remains naive repeated sampling: draw many independent solutions and hope one is correct. Because such sampling explores only through local decoding noise, it tends to produce many near duplicate attempts rather than genuinely different ideas. We ask whether exploration can instead be steered at a semantic level, by first sampling problem specific concepts, hints, or strategies and then conditioning answer generation on them. We refine this into a simple, more exploratory procedure that emits many diverse concepts in a single trajectory, and evaluate it on hard problems where repeated sampling struggles.
We then go a step further and make concept generation \emph{trainable}: a small concept generator is optimized with reinforcement learning so that its concepts maximize the downstream success of a larger, frozen answer generator. On hard mathematical reasoning problems, the trained concept generator substantially improves the answer generator's pass@k over naive repeated sampling at the same answer generation allocation, surpasses concepts drawn from much larger untuned models, and transfers to answer generators it was never trained against, including a model from a different family. A small model can thus be trained into an effective, reusable search policy for a much larger one.

}

\correspondence{\email{ilabiad@meta.com}}

\begin{document}

\maketitle

\section{Introduction}\label{sec:introduction}
Reinforcement learning has become a central ingredient in training modern large language models (LLMs), particularly for reasoning: a model is improved by sampling its own solutions, scoring them against a verifier or reward, and reinforcing the successful ones \citep{shao2024deepseekmath,olmo2025olmo}. The effectiveness of this loop hinges on exploration. RL can only reinforce, and ultimately distill into the model weights, behaviors that the sampling policy manages to discover in the first place. On a hard problem where the model never stumbles on a correct solution, there is simply no positive signal to learn from. The same bottleneck appears at inference time, where discovering a correct solution among many samples is precisely what test-time scaling relies on \citep{snell2024scaling}. In both regimes, progress is limited by the quality of the underlying search policy, which makes the ability to explore the solution space effectively a first class concern rather than an implementation~detail.

In practice, the dominant form of exploration is repeated sampling: draw many complete answers from the same model and keep those a verifier or judge deems correct \citep{brown2024large,wang2022self}. This explores only indirectly, through local token level decoding noise, and often yields many near duplicate attempts rather than genuinely different ideas. A natural alternative is to separate deciding how to approach a problem from carrying out the solution: first sample high level concepts, hints, theorems, or strategies, and then generate answers conditioned on them. This mirrors how people tackle hard problems, pausing to consider which ideas or tools might apply before committing to a full derivation, diversifying reasoning at a semantic level rather than at the level of individual tokens.

Recent work instantiated this idea as a concept guided sampling pipeline, using one LLM to propose concepts and another to answer conditioned on them \citep{handa2026guidedsampling}. In our reproduction, however, its advantage largely vanishes once the repeated sampling baseline is allowed to use exploratory decoding parameters instead of the restrictive ones originally used. We further identify two practical limitations: concepts are produced one at a time in an iterative loop, which is wasteful given that LLMs are autoregressive, and only about one concept is generated per problem on average, sharply limiting the diversity the method was meant to provide.

In this paper we revisit concept guided exploration for mathematical reasoning and turn it into a trainable component. We first strengthen the inference time procedure: we generate all concepts in a single trajectory, prompt for detailed and problem specific concepts, and evaluate on hard subsets where naive repeated sampling fails outright. Building on this, we ask whether a small concept generator can be \emph{trained}, rather than merely prompted, to produce concepts that improve a larger, frozen answer generator that may be impractical to train, or even closed-source. Concretely, we cast exploration itself as an RL problem, but over concepts rather than answers: we train the concept generator with reinforcement learning, rewarding each concept trajectory by the downstream success of the answer generator it steers. On a held-out set of hard problems, this doubles the answer generator's pass@128, from $19.0\%$ for naive repeated sampling to $39.2\%$ at the same answer generation allocation. Moreover, the learned policy behaves as a reusable search strategy rather than a scale effect: a trained 7B concept generator steers a 32B answer generator well enough to surpass the concepts that larger model produces for itself, and it transfers to a 70B answer generator from a different family that it was never trained against. Our contributions are:
\begin{itemize}
  \item Re-evaluating GuidedSampling across all 25 Qwen2.5 CG/AG size pairs (1.5B--32B) on MATH500, we find that its gains largely disappear against an exploration-enabled repeated sampling baseline.
  \item We introduce single trajectory concept generation and RL-train Qwen2.5-7B as a search policy for a frozen Qwen2.5-32B answer generator, raising pass@128 from $19.0\%$ to $39.2\%$ on 1k held-out hard DeepMath problems.
  \item The trained 7B CG outperforms an untuned 32B CG and transfers without retraining to Llama-3.3-70B; its gains also persist out of distribution on Omni-MATH~2.
  \item We analyze whether the learned generator improves exploration rather than merely solving problems directly, and study answer leakage, concept relevance, position effects, and the effect of the number of concepts.
\end{itemize}

\section{Related Work}\label{sec:relatedwork}
\paragraph{Inference-Time Search and Concept-Guided Exploration}
Repeated sampling is a standard way to spend inference-time compute on reasoning, with performance scaling in the number of samples \citep{brown2024large,snell2024scaling} and typically paired with self-consistency or verifier-based selection \citep{wang2022self,lightman2024let}. A complementary line of work decomposes reasoning into intermediate artifacts (plans, hints, or solution sketches) before producing an answer \citep{wang2025planning}. Our direct predecessor is the GuidedSampling method of \citet{handa2026guidedsampling}, which samples concepts from one LLM and conditions a second LLM's answers on them; unlike them, we find in our reproduction that this advantage largely disappears once the repeated-sampling baseline is allowed to explore (\cref{sec:background}), and we depart from their setup by emitting all concepts in a single trajectory and by training the concept generator for downstream answer generator success rather than using it zero-shot. This also contrasts with tree-search methods such as Tree of Thoughts \citep{yao2023tree}, whose dynamic branching disrupts continuous batching and KV-cache sharing in optimized engines like vLLM \citep{kwon2023efficient}, whereas our single trajectory generation diversifies exploration while preserving standard autoregressive throughput.

\paragraph{RL-Driven Abstractions and Small-Model Orchestration}
The closest RL-based method is RLAD \citep{qu2025rlad}, which trains an abstraction generator to produce procedural summaries that guide a downstream solver. Unlike RLAD, our method uses no SFT warm start based on traces from a stronger teacher, targets concise, problem-specific exploration hints rather than full procedural summaries, and trains only the concept generator.
Like SOAR \citep{sundaram2026teaching}, we reward the generator using measured downstream performance rather than an intrinsic proxy, however, our downstream model remains frozen.
Related zero-shot methods elicit similar artifacts without training the generator: Step-Back prompting abstracts a problem to higher-level principles \citep{zheng2024take}, and Auto-CoT synthesizes its own few-shot exemplars \citep{zhang2022automatic} (whose gains may stem from problem creation rather than the examples themselves \citep{gwak2026not}). In contrast, we \emph{train} the concept generator with trajectory-level rewards from downstream answer success. Finally, our setup echoes small-model orchestration of larger systems: Directional Stimulus Prompting \citep{li2023guiding} trains a small LM to emit hints that guide a frozen black-box LLM, and ToolOrchestra \citep{su2025toolorchestra} RL-trains a lightweight orchestrator to coordinate tools and frontier models for efficiency, whereas we steer a single fixed, larger answer generator through concepts, with a trained 7B concept generator improving a 32B answer generator.

\section{Background and Analysis}\label{sec:background}
The method most closely related to ours first samples concepts with one LLM and then uses a second LLM to generate final answers conditioned on those concepts \citep{handa2026guidedsampling}. The intended benefit is to explore diverse mathematical ideas before spending the larger answer-generation allocation. In principle, this can produce more semantically diverse rollouts than repeated sampling from the answer generator alone.

\textbf{Reproduction and Baseline Sensitivity.}
We reproduce concept-guided sampling on MATH500~\citep{hendrycks2021measuring,lightman2024let}, using Qwen2.5-Instruct models at five sizes ($1.5$B, $3$B, $7$B, $14$B, and $32$B) as both the concept generator (CG) and the answer generator (AG), and evaluate all $25$ CG/AG size combinations. Concepts are produced with the original iterative generation procedure and prompts from \citet{handa2026guidedsampling}, left unchanged. For each problem we draw a total of $100$ answer rollouts and report pass@$50$: the baseline draws all $100$ rollouts directly from the AG, whereas the concept-guided condition splits the same $100$-rollout allocation across the concepts parsed from the CG trajectory, so the two conditions are matched in answer-generation compute. Only the AG sampling parameters differ from the original protocol.

The reported advantage of concept-guided sampling is highly sensitive to these parameters. The original repeated-sampling baseline used temperature $0.8$ and top-$p=0.5$, which restricts exploration and yields an artificially weak baseline. When both the baseline and the concept-conditioned answer generator are instead sampled with exploratory parameters (temperature $1.0$, top-$p=0.95$, top-$k$ disabled, implemented as $-1$), the concept-guided method no longer improves over repeated sampling and slightly degrades it across nearly all CG/AG pairs. \Cref{tab:guided_sampling_reproduction} reports, for each AG size, the baseline pass@50 (top row) and the pass@50 difference between the concept-guided method and this baseline for every CG (rows) / AG (columns) pair, where negative values indicate underperformance; the final column gives the average number of concepts generated per problem by each CG. Once the baseline is allowed to explore, the gains attributed to concept guidance largely disappear.

\begin{table}[!ht]
\centering
\caption{\label{tab:guided_sampling_reproduction}Reproduction of concept-guided sampling \citep{handa2026guidedsampling} on MATH500 (Qwen2.5-Instruct): baseline pass@50 and concept-guided differences across CG/AG size pairs. Reported gains disappear once the answer generator is allowed to explore freely.}
\begin{tabular}{lcccccc}
\toprule
& \multicolumn{5}{c}{AG} & \multirow{2}{*}{\shortstack{Avg\\Concepts}}\\
\cmidrule(lr){2-6}
& 1.5B & 3B & 7B & 14B & 32B & \\
\midrule
Baseline pass@50 & 89.2 & 93.8 & 95.9 & 96.0 & 96.2 & - \\
\addlinespace
\midrule
\textbf{CG / AG} &  \multicolumn{5}{l}{\textbf{Difference (Concept - Baseline) pass@50}}  \\
\addlinespace
\quad 1.5B & -0.6 & -0.7 & -1.0 & -0.8 & -0.8 & 1.15 \\
\quad 3B & -1.7 & -0.6 & -1.3 & -0.8 & -1.0 & 1.17 \\
\quad 7B & -1.8 & +0.1 & -1.8 & -1.2 & -0.7 & 1.02 \\
\quad 14B & -1.0 & -0.7 & -2.2 & -1.1 & -1.8 & 1.25 \\
\quad 32B & -1.4 & -0.1 & -0.8 & -0.2 & -0.5 & 1.68 \\
\bottomrule
\end{tabular}
\end{table}

We also identify two structural limitations. First, concepts are generated one by one in an iterative loop. This is slow and unnecessary because LLMs are autoregressive and will generate newer concepts conditioned on the older ones. It is then possible to generate a list of diverse concepts in a single trajectory.
Second, the method generates approximately one concept per problem on average (\cref{tab:guided_sampling_reproduction}), which defeats the purpose of exploring multiple solution ideas. This low concept count is consistent with the original work, which observes that Qwen2.5 models generate noticeably fewer concepts than their Llama-3.2 counterparts and reports only $1.13$ distinct concepts per problem on HumanEval~\citep{handa2026guidedsampling}.
The disappearance of gains is not merely an artifact of this low concept count. Llama-3.2-3B-Instruct generates $3.18$ concepts per problem on MATH500, 3 times more than the Qwen models, yet under exploratory sampling concept guidance still does not help: it reaches $86.7\%$ pass@50 versus $88.1\%$ for the repeated-sampling baseline (a $-1.4$ point change). Even when the concept generator produces multiple concepts, allowing the baseline to explore erases the reported advantage, indicating that the baseline sampling parameters, not the number of concepts, drive the effect.

These observations motivate our concept-guided sampling procedure in \cref{sec:concept_guided_sampling}.

\section{Concept-Guided Sampling}\label{sec:concept_guided_sampling}
We first modify concept-guided sampling as an inference-time procedure. The goal is not to train a model yet, but to test whether a better concept-generation protocol can improve exploration beyond naive repeated sampling on problems where the answer generator struggles.

\textbf{Exploratory Answer Generation}.
We keep the exploratory decoding parameters introduced in \cref{sec:background} (temperature $1.0$, top-$p=0.95$, and top-$k=-1$) for concept-conditioned generation and repeated sampling. This keeps the comparison conservative: any improvement from concepts must exceed what a strong, exploration-enabled repeated-sampling baseline already achieves.

\textbf{Single-Trajectory Concept Generation}.
Instead of sampling concepts one by one, we prompt the concept generator to analyze the problem and emit all useful concepts in one trajectory. Each concept is required to be high-signal, problem-specific, and non-duplicative. We parse at most ten concepts from the trajectory and condition answer rollouts on those concepts. The full concept-generator and answer-generator prompts are given in \cref{app:cg_prompt,app:ag_prompt}.

\textbf{Focusing on Hard Problems}.
On easy problems, repeated sampling already solves most examples, leaving little room for concept-guided exploration to help. We therefore construct model specific hard subsets from the MATH test set: for each answer-generator size, we retain only problems for which the repeated-sampling baseline obtains $0\%$ accuracy with 100 rollouts. Each answer generator is thus evaluated on its own filtered subset. The hard subsets shrink as the answer generator grows, from $493$ problems for the $1.5$B model to $178$ for the $32$B model, since more capable models leave fewer completely unsolved problems.

\begin{table}[!ht]
\centering
\caption{\label{tab:concepts_ours}Our concept-guided sampling (without training), on model-specific hard MATH subsets (Qwen2.5-Instruct): baseline pass@50 and concept-guided differences across CG/AG size pairs. The final row reports the size of each answer generator's hard subset.}
\begin{tabular}{lcccccc}
\toprule
& \multicolumn{5}{c}{AG} & \multirow{2}{*}{\shortstack{Avg\\Concepts}}\\
\cmidrule(lr){2-6}
& 1.5B & 3B & 7B & 14B & 32B & \\
\midrule
Baseline pass@50 & 18.4 & 16.6 & 11.0 & 12.6 & 11.8 & - \\
\midrule
\addlinespace
\textbf{CG / AG} &  \multicolumn{5}{l}{\textbf{Difference (Concept - Baseline) pass@50}} \\
\addlinespace
\quad 1.5B & -1.4 & -1.0 & +1.7 & +6.0 & +3.3 & 4.18 \\
\quad 3B & +0.1 & +0.4 & +8.3 & +4.4 & +8.7 & 8.74 \\
\quad 7B & +1.7 & +2.1 & +8.0 & +3.5 & +8.8 & 9.75 \\
\quad 14B & -0.2 & +5.2 & +7.2 & +7.4 & +8.2 & 9.95 \\
\quad 32B & +3.3 & +3.8 & +9.7 & +7.4 & +6.6 & 9.95 \\
\midrule
Dataset size & 493 & 274 & 185 & 176 & 178 & - \\
\bottomrule
\end{tabular}
\end{table}

\textbf{Inference time results.}
In contrast to the reproduction in \cref{sec:background}, where concept guidance never helped once the baseline was allowed to explore, our procedure produces consistent gains on the hard subsets (\cref{tab:concepts_ours}). Almost every concept-generator (CG) / answer-generator (AG) pair improves over the repeated-sampling baseline, with gains of up to $+9.7$ pass@50 points. The improvements scale with the strength of the concept generator: the $1.5$B generator helps only marginally, whereas generators of $7$B and larger improve nearly every answer generator by several points. Running the iterative protocol~\citep{handa2026guidedsampling} on these same hard subsets with the same AG sampling parameters also yields gains, but smaller ones (at most $+6.8$ points, with several regressions), consistent with its far lower concept count (\cref{app:guided_sampling_hard}).

Most importantly, a small concept generator can improve a much larger answer generator. The $7$B generator raises the $32$B answer generator by $+8.8$ points despite being far smaller than the model it steers, indicating that useful exploration can be produced with little compute and injected into a stronger, possibly frozen or closed-source, answer generator. This observation directly motivates the training setup in \cref{sec:training_concept_generators}, where a small concept generator is trained for a larger, fixed answer generator. An example of generated concepts with our approach is given in \cref{ex:concepts}.

These gains coincide with a large increase in the number of concepts. Whereas the original procedure produced about one concept per problem (\cref{tab:guided_sampling_reproduction}), single-trajectory generation yields between $4$ and roughly $10$ concepts on average, with the stronger generators saturating the cap of ten.

\begin{examplebox}[ex:concepts]{A MATH problem and example concepts}
\textbf{Problem:}\\
A bag contains two red beads and two green beads. You reach into the bag and pull out a bead, replacing it with a red bead regardless of the color you pulled out.  What is the probability that all beads in the bag are red after three such replacements? Express your answer as a common fraction.

\medskip
\textbf{Example concepts:}
\begin{itemize}
  \item Use a tree diagram to map all possible outcomes of the draws and replacements.
  \item Calculate the probability for each state transition step-by-step.
  \item Analyze the problem using a Markov chain to model the transitions between states.
\end{itemize}
\end{examplebox}

\textbf{Cross-family check.} Our reported gains are not specific to the Qwen family. On a hard subset of $424$ MATH problems that Llama-3.2-3B-Instruct fails to solve with $100$ rollouts, our single-trajectory procedure (using Llama-3.2-3B as both concept and answer generator) yields a modest but positive improvement, raising pass@50 from $25.2\%$ to $26.1\%$ ($+0.9$ points), while producing $9.23$ concepts per problem on average, far more than the one to three produced by the original iterative procedure of \citet{handa2026guidedsampling}; see \cref{sec:background}.
Notably, that original procedure \emph{degraded} this same model's exploratory baseline; switching to single-trajectory generation both increases concept diversity and flips the effect positive, confirming the improvement is not an artifact of the Qwen family.

\section{Training Concept Generators}\label{sec:training_concept_generators}
\begin{figure}[t]
  \centering
  \includegraphics[width=\linewidth]{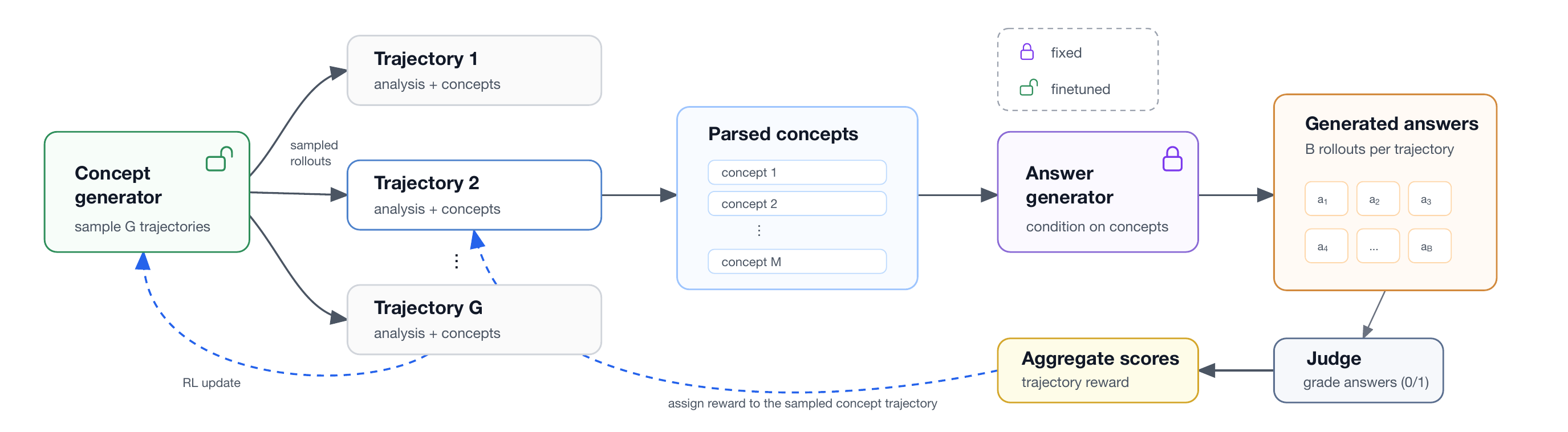}
  \caption{Overview of concept-generator training. The concept generator samples multiple concept trajectories. For one sampled trajectory, parsed concepts condition a fixed answer generator, which produces answer rollouts. A judge grades the answers, and the scores are aggregated into a trajectory-level reward used to finetune the concept generator.}
  \label{fig:method_overview}
\end{figure}

The inference time results raise a natural question: can we {\em train} a small model to generate better concepts for a larger answer generator? This is useful because the larger answer generator may be impractical to train directly or its weights unavailable. In particular, this provides a ``training handle'' for otherwise untrainable models, including those only accessible via API. Throughout, the answer generator remains frozen and only the concept generator is trained.
\Cref{fig:method_overview} summarizes the full training loop, and \cref{app:theory} gives the formal setting with a fuller discussion of the reward~design.

\textbf{Concept sampling and answer rollouts}.
Let $x$ denote an input problem and $\pi_\theta$ the concept generator (CG); the answer generator (AG) is a fixed model used only for inference. For each problem, $\pi_\theta$ samples $G$ trajectories. Each trajectory contains a problem analysis followed by a list of concepts. We parse the generated text and keep at most $10$ concepts. For each concept trajectory, we assign a total answer-rollout allocation of $B$ calls to the answer generator (AG). The allocation is distributed as evenly as possible across the parsed concepts: if $M$ concepts are available, each concept receives either $\lfloor B/M \rfloor$ or $\lceil B/M \rceil$ rollouts, with the total number of rollouts equal to $B$. This handles cases such as $B=128$ and $M=10$, where some concepts receive $13$ rollouts and the rest receive $12$.

\textbf{Reward aggregation}.
Each AG rollout is graded for correctness against the ground-truth answer using an LLM judge.
We use an LLM judge because, even after extracting the final answer, rule-based matching can penalize mathematically equivalent answers expressed in different forms or formats.
We then aggregate the rollout scores into a scalar reward for the entire concept trajectory. Consider a trajectory $\tau$ with parsed concepts $c_1,\dots,c_M$, where concept $c_j$ receives $n_j$ answer rollouts with binary correctness $o_{j,\ell}\in\{0,1\}$. We consider two aggregation functions,
\begin{equation}
r_{\text{max-max}}(\tau)=\max_{j}\max_{\ell}\, o_{j,\ell},
\qquad
r_{\text{max-mean}}(\tau)=\max_{j}\frac{1}{n_j}\sum_{\ell=1}^{n_j} o_{j,\ell}.
\end{equation}
\textbf{Max-of-max} assigns reward $1$ if any answer rollout generated from any concept in the trajectory is correct, and $0$ otherwise; it rewards discovering at least one concept that unlocks a correct solution. \textbf{Max-of-mean} takes the maximum over concepts of the per-concept accuracy; it rewards concepts that are reliably useful rather than concepts that only occasionally lead to a correct sample. The two aggregations therefore differ in range: max-of-max yields a binary reward in $\{0,1\}$, whereas max-of-mean yields a continuous reward in $[0,1]$.

\textbf{Trajectory-level credit}.
To encourage a diverse set of concepts, we assign the aggregate reward to the entire concept trajectory instead of attributing reward to individual concepts. This encourages exploration over sets of concepts, but it may introduce position effects or allow weak concepts to share credit with strong ones. We study these effects in~\cref{sec:pos_biais}.

\textbf{Optimization}.
We train the CG policy $\pi_\theta$ with a GRPO-style objective \citep{shao2024deepseekmath}. The full objective and optimization details are provided in \cref{sec:experiments} and \cref{app:training_details}.

\section{Experiments}\label{sec:experiments}
\subsection{Setup}
We now test our approach introduced in the previous section to train the CG model.

\textbf{Models.}
We use Qwen2.5-7B-Instruct as the trainable concept generator and Qwen2.5-32B-Instruct as the primary answer generator \citep{qwen2.5}. The answer generator is kept fixed and used solely for inference. The transfer experiments in \cref{sec:ablations} further use Llama-3.3-70B-Instruct \citep{grattafiori2024llama} as an answer generator.

\textbf{Data.}
Training uses problems and ground truths from DeepMath-103k \citep{he2026deepmath}. We filter the training set using the answer generator and retain problems where the AG achieves less than $5\%$ success with 128 rollouts. As the held-out evaluation set we reserve $1$k problems, disjoint from training, where the AG achieves $0\%$ success with 128 rollouts; these are the hardest problems for the answer generator and are used for all main results. Because this filter is a finite-sample estimate, a fresh independent draw can still solve a fraction of these problems; we quantify this effect in \cref{app:filtering_variance}. To test generalization beyond the training distribution, we additionally evaluate on Omni-MATH~2 \citep{ballon2026benchmarkssaturatewhen}, itself a filtered version of Omni-MATH \citep{gao2025omni}, which we filter with the exact same procedure (retaining problems where the AG achieves $0\%$ success with 128 rollouts) to obtain an out-of-distribution hard set. We do not train on this dataset; it is used only for evaluation, following the same protocol as the DeepMath held-out set.

\textbf{Training Algorithm.}
We train the CG with GRPO-style reinforcement learning, using the variant adopted in OLMo3 \citep{olmo2025olmo}. For each problem, the CG samples $G=8$ concept trajectories. Each trajectory receives a downstream reward computed from $B=128$ AG rollouts. Because computing each trajectory reward requires many AG rollouts, we do not use dynamic/active sampling during training. We train for 200 batches and report both reward-aggregation variants defined in \cref{sec:training_concept_generators} (max-of-max and max-of-mean). Detailed hyperparameters are reported in~\cref{app:training_details}.

\textbf{Answer Generation and Judging.}
The AG uses temperature $1.0$, top-$p=0.95$, top-$k=-1$, a maximum response length of 4096 tokens, and a maximum prompt length of 2048 tokens. We judge answer correctness with Qwen2.5-32B-Instruct using greedy decoding and a binary verifier prompt. We deliberately reuse the answer-generator model as the judge for efficiency: a separate high-quality judge would need to be comparably large and would consume dedicated GPUs, slowing an already compute-heavy training loop. The judge is invoked in a separate scoring prompt that compares a candidate answer after extraction to the ground truth. The judge prompt is given in \cref{app:judge_prompt}.

\textbf{Baselines.}
We compare:
\begin{itemize}
  \item \textbf{Naive repeated sampling:} sample answers directly from the AG with the same rollout allocation.
  \item \textbf{Generic prompt modification:} sample answers using the concept conditioned AG prompt, using ten fixed, problem-agnostic hints (e.g.\ ``break the problem into smaller subproblems''), with the rollout allocation split across them exactly as in concept guided sampling. The full list of hints is given in \cref{app:generic_prompt}.
  \item \textbf{Untuned concept generation (Ours untrained):} use pretrained concept generators, including larger models, without RL training, as in the inference-time procedure of \cref{sec:concept_guided_sampling}.
  \item \textbf{Trained concept generation (Ours trained):} the RL-trained concept generator steering the fixed answer generator.
\end{itemize}

\textbf{Metric.}
The primary metric is pass@k as a function of the answer-rollout allocation $k$, computed with the standard unbiased estimator of \citet{chen2021evaluating}. Unless otherwise stated, curves are evaluated on the 1k DeepMath held-out set of hard problems.

\subsection{Results}

\begin{table}[!ht]
\centering
\small
\setlength{\tabcolsep}{4pt}
\caption{\label{tab:main_results}Pass@$k$ (\%) on DeepMath held-out hard set and Omni-MATH~2 OOD hard set. AG is Qwen2.5-32B-Instruct, CG is Qwen2.5-7B-Instruct. Best per column in \textbf{bold}, second best in \emph{italic}.}
\begin{tabular}{lrrrrrr}
\toprule
& \multicolumn{3}{c}{DeepMath (held-out hard)} & \multicolumn{3}{c}{Omni-MATH~2 (OOD)} \\
\cmidrule(lr){2-4} \cmidrule(lr){5-7}
Method & pass@1 & pass@64 & pass@128 & pass@1 & pass@64 & pass@128 \\
\midrule
\multicolumn{7}{l}{\emph{Baselines}}\\
\quad Naive repeated sampling            & 0.23 & 11.35 & 19.00 & 0.14 & 6.81 & 11.29 \\
\quad Generic prompt modification        & 0.32 & 14.54 & 22.60 & 0.15 & 6.93 & 11.22 \\
\addlinespace
\multicolumn{7}{l}{\emph{Untuned concept generation (ours, no RL training)}}\\
\quad Untuned CG (32B)                    & 0.82 & 23.77 & 33.80 & 0.27 & \textit{10.18} & \textit{15.27} \\
\quad Untuned CG (7B)                     & 0.77 & 20.62 & 28.90 & \textit{0.28} & 9.64 & 14.98 \\
\addlinespace
\multicolumn{7}{l}{\emph{Trained concept generation (ours, 7B CG)}}\\
\quad max-of-max                          & \textit{1.48} & \textit{26.66} & \textit{35.30} & 0.27 & 10.04 & \textit{15.27} \\
\quad max-of-mean                         & \textbf{2.18} & \textbf{29.64} & \textbf{39.20} & \textbf{0.44} & \textbf{12.75} & \textbf{18.60} \\
\bottomrule
\end{tabular}
\end{table}

\textbf{Main Results.}
The trained concept generators substantially improve pass@k over naive repeated sampling on the held-out hard problems (\cref{tab:main_results}). The best variant (max-of-mean) reaches $39.2\%$ pass@128, roughly double the $19.0\%$ of naive repeated sampling, and the gap holds at smaller allocations ($29.6\%$ vs.\ $11.4\%$ at pass@64). A generic, problem-agnostic prompt modification helps only modestly ($22.6\%$ pass@128), far below problem-specific concepts. Both aggregation objectives improve over the baselines, with max-of-mean ahead of max-of-max at every allocation ($39.2\%$ vs.\ $35.3\%$ pass@128). This indicates the CG learns to propose problem-specific directions that reshape the AG's exploration distribution rather than merely drawing more samples from the same distribution. We verify this exploration effect directly: embedding the generated chains of thought and measuring their diversity with the Vendi Score shows that concept guidance, and especially the trained CG, produces more diverse reasoning than naive sampling (\cref{app:diversity}). The improvement carries over to the out-of-distribution Omni-MATH~2 set, where max-of-mean is again the best method ($18.6\%$ vs.\ $11.3\%$ pass@128 for naive), though margins are smaller and max-of-max there only matches the untrained concept generators. The main table reports the headline allocations $k\in\{1,64,128\}$; the full pass@k curves for all methods and both datasets, together with 95\% bootstrap confidence intervals and paired-bootstrap tests of the gains over naive sampling, are reported in \cref{app:extended_curves}. We train for $200$ steps throughout; \cref{app:more_steps} shows that training substantially longer yields no sustained gain: performance saturates after a few hundred steps, and the max-of-mean objective eventually becomes unstable, which justifies this choice.

\textbf{Small Concept Generators Can Help Larger Answer Generators.}
A central finding is that a small \emph{trained} CG outperforms a much larger \emph{untrained} one. Without training, a larger concept generator does help more, consistent with scale: the untuned 32B CG reaches $33.8\%$ pass@128 versus $28.9\%$ for the untuned 7B CG. After RL training, however, the 7B CG surpasses the untuned 32B CG ($39.2\%$ vs.\ $33.8\%$ pass@128 for max-of-mean, and $35.3\%$ for max-of-max), despite steering an answer generator more than four times its size. This supports the view that the concept generator learns a specialized search policy rather than relying on model scale, and that useful exploration can be produced with little compute and injected into a larger frozen answer generator. A paired bootstrap over held-out problems confirms this gap: $+5.4$ points pass@128 on DeepMath and $+3.3$ on Omni-MATH~2, with both $95\%$ intervals excluding $0$ (\cref{tab:passk_ci_pairwise} in \cref{app:extended_curves}).

\textbf{Comparison to Inference-Time Concept Sampling.}
The untuned concept generators in \cref{tab:main_results} are exactly the inference-time procedure of \cref{sec:concept_guided_sampling}, and they already improve over naive sampling ($28.9\%$ pass@128 for the 7B CG). RL training adds a further $+10.3$ points over the untuned CG, showing that training improves concept quality. \cref{app:concept_examples} shows qualitative examples of how training reshapes generated concepts into more concrete, problem-specific plans. We confirm this improvement is statistically significant with a paired bootstrap over held-out problems (per-problem difference of the two unbiased pass@k estimates); \cref{tab:passk_ci_vs7b} in \cref{app:extended_curves} reports the trained over untuned 7B gain for both reward objectives and datasets.

\section{Analysis and Ablations}\label{sec:ablations}
\subsection{Do Problem-Relevant Concepts Matter?}
We test whether the gains come from problem-specific guidance or merely from perturbing the prompt. Using the trained CG concepts (max-of-mean), we apply a random derangement over the held-out set so that each problem is paired with another problem's concept trajectory (no problem keeps its own), and re-run answer generation.
Mismatched concepts still help slightly over naive sampling ($23.9\%$ vs.\ $19.0\%$ pass@128), on par with the generic prompt-modification baseline ($22.6\%$ pass@128), consistent with a concept from an unrelated problem acting as a generic perturbation. But they fall far short of matched concepts ($39.2\%$ pass@128): problem relevance accounts for roughly $15$ of the $20$ point improvement over naive sampling. This indicates that most of the benefit comes from problem-specific concepts, not from generic prompt perturbation.

\subsection{Transfer Across Datasets and Models}
\label{sec:transfer}
\begin{table}[!ht]
\centering
\small
\setlength{\tabcolsep}{4pt}
\caption{\label{tab:transfer}Transfer to Llama-3.3-70B as the answer generator, on the DeepMath and Omni-MATH~2 hard sets. Pass@$k$ (\%); best per column in \textbf{bold}, second best in \emph{italic}.}
\begin{tabular}{lrrrrrr}
\toprule
& \multicolumn{3}{c}{DeepMath (held-out hard)} & \multicolumn{3}{c}{Omni-MATH~2 (OOD)} \\
\cmidrule(lr){2-4}\cmidrule(lr){5-7}
Method (AG = Llama-3.3-70B) & pass@1 & pass@64 & pass@128 & pass@1 & pass@64 & pass@128 \\
\midrule
Naive repeated sampling            & \textit{6.17} & 22.38 & 26.20 & \textbf{1.92} & 9.63 & 12.50 \\
\addlinespace
\multicolumn{7}{l}{\emph{Untuned concept generation (no RL training)}}\\
\quad Llama-70B self-concepts      & 6.15 & 24.91 & 28.90 & \textit{1.88} & 9.77 & 11.94 \\
\quad Untuned 7B CG (Qwen)         & 5.66 & \textit{26.24} & \textit{31.30} & 1.73 & \textit{9.92} & \textit{12.59} \\
\addlinespace
\multicolumn{7}{l}{\emph{Trained concept generation (7B CG, max-of-mean)}}\\
\quad 7B CG trained vs.\ 32B (transfer) & \textbf{6.83} & \textbf{29.01} & \textbf{34.30} & 1.73 & \textbf{10.75} & \textbf{13.97} \\
\bottomrule
\end{tabular}
\end{table}

\textbf{Cross-model transfer.}
Our concept generator is trained only against the Qwen2.5-32B-Instruct answer generator and never sees Llama-3.3-70B-Instruct \citep{grattafiori2024llama} during training. To test whether the learned search policy transfers, we take the trained 7B CG and use it to steer Llama-3.3-70B, a larger model from a different family, on both the 1k DeepMath held-out set and the Omni-MATH~2 set. We reuse both filtered sets exactly as in \cref{sec:experiments} (filtered so the Qwen2.5-32B AG scores $0\%$, and reused as-is rather than re-filtered for Llama), so they measure transfer on common hard sets rather than Llama's own worst-case problems. Accordingly the Llama baseline is non-zero, solving $6.2\%$ of the DeepMath problems at pass@1. For all rows we judge correctness with Qwen2.5-32B-Instruct, consistent with our earlier experiments.
As shown in \cref{tab:transfer}, on DeepMath the transferred CG is best at every $k$ value, reaching $34.3\%$ pass@128 versus $26.2\%$ for naive sampling and $28.9\%$ for Llama generating its \emph{own} concepts. The small trained CG thus outperforms the $10\times$larger answer generator's self-generated concepts, despite never being trained against it.
This indicates the CG learns a transferable exploration policy rather than one tied to a specific AG.

\subsection{Training and Inference Compute}
The training loop of our approach (\cref{sec:training_concept_generators}) is compute-heavy, and this compute is almost entirely dominated by reward estimation. Before a single gradient is taken, scoring the $G=8$ concept trajectories for a problem requires $G\times B=1024$ generations from the frozen answer generator plus $1024$ judge calls per problem in the batch. In a representative training step of about $413$\,s, the AG rollouts take $\sim\!300$\,s and judging $\sim\!85$\,s (together roughly $94\%$ of the step) while generating the concept trajectories with the CG takes only $\sim\!18$\,s and the actor update about $10$\,s. Training is therefore bottlenecked by repeatedly running the large model we do \emph{not} train, not by the small concept generator we optimize. Crucially, training is run only once.

At inference, running the CG to sample one concept trajectory per problem adds negligible compute next to running $B$ AG rollouts per problem. Using the standard estimate of $2N$~\citep{kaplan2020scaling,snell2024scaling} FLOPs per token for a dense transformer with $N$ parameters, a single generation requires about $2N(\ell_{\text{in}}+\ell_{\text{out}})$ FLOPs, so the compute of one CG call relative to one AG rollout is roughly $\tfrac{N_{\text{CG}}}{N_{\text{AG}}}\cdot\tfrac{\ell_{\text{CG}}}{\ell_{\text{AG}}}$. With $N_{\text{CG}}=7.6$B and $N_{\text{AG}}=32.5$B parameters and average lengths of $\ell_{\text{CG}}\approx280+800=1080$ and $\ell_{\text{AG}}\approx240+820=1060$ tokens, this ratio is about $\tfrac{7.6}{32.5}\cdot\tfrac{1080}{1060}\approx0.24$: generating the concepts takes less than a quarter of a single answer rollout. Furthermore, it is under $0.3\%$ of the $B=128$ rollout answer allocation (we ignore $\ell_{\text{in}}$ for the AG due to prefix sharing). Therefore this allocation is essentially the \emph{same} as naive repeated sampling. We do not add answer rollouts, we only prepend one lightweight small-model call and change how the existing $B$ rollouts are prompted, so concept-guided exploration runs at essentially the same inference compute as the~ baseline.

\subsection{Additional Ablations}
We report further analyses in \cref{app:additional_results}.
Training the 7B model to answer directly with the same RL algorithm, training split, and judge performs worse than using it as a concept generator for the frozen 32B answer generator, showing that the gain comes from steering a stronger model rather than teaching the small model to solve the problems itself (\cref{app:direct_training}).
Per-concept accuracy remains within a narrow range across trajectory positions, and the bootstrap confidence intervals for positions~2--10 broadly overlap: no position is dead, supporting our use of trajectory-level credit assignment (\cref{sec:pos_biais}).
Answer leakage is rare as only $0.44\%$ of concepts are flagged as containing the ground-truth answer, and only $3\%$ of problems contain any flagged concept; this proves that the gains cannot be explained by answer leakage (\cref{app:answer_leakage}).
Finally, when the total $B=128$ answer-rollout budget is held fixed, increasing the number of concepts from one to ten raises pass@$128$ from $22.7\%$ to $39.8\%$ while pass@$1$ remains nearly unchanged, showing that distributing a fixed rollout budget across more concepts improves coverage of the solution space, especially at larger $k$ (\cref{app:num_concepts}).

\section{Conclusion}\label{sec:conclusion}
We revisited LLM exploration beyond naive repeated sampling. After showing that prior concept-guided gains largely disappear against an exploratory baseline, we introduced a stronger procedure that generates multiple detailed concepts in one trajectory and trains a small concept generator using the downstream success of a frozen answer generator. On hard mathematical problems, the learned policy roughly doubles pass@k at the same answer-rollout allocation, outperforms concepts from much larger untuned models, and transfers to an answer generator from a different model family. These results show that a small model can serve as an effective, reusable search policy for a larger model that may be impractical to train directly or closed-source.

Several limitations remain. Training is compute heavy because reward estimation requires many answer-generator rollouts, rewards are assigned at the trajectory level rather than to individual concepts, and our evaluation focuses on mathematical reasoning. Future work should develop lower-compute reward estimation, more granular credit assignment, judge-robust evaluation, and bandit based allocation of rollouts across concepts, while extending the approach to broader domains.

\bibliographystyle{plainnat}
\bibliography{references}

\appendix
\crefalias{section}{appendix}
\crefalias{subsection}{appendix}
\crefalias{subsubsection}{appendix}
\clearpage

\section{Formal Setting and Analysis of the Reward Objectives}\label{app:theory}
\paragraph{Objective.} Our goal is exploration: finding at least one valid solution path within a sampling allocation $k$. This is exactly the definition of $\text{pass}@k$, which we report per problem $x$ and average over the dataset.

Let $q(\cdot\,|\,x)$ denote the \emph{frozen} model that generates candidate answers, and let $\pi_\theta(\cdot\,|\,x)$ denote the \emph{trained} policy, which generates concepts. Write \[ p(x) \;=\; \mathbb{E}_{y\sim q(\cdot|x)}\big[\mathbbm{1}\{y\text{ correct}\}\big] \] for the probability that naive sampling from the frozen model produces a correct answer. Throughout we assume (i) rollouts are drawn i.i.d.\ from the relevant conditional of $q$, and (ii) the verifier is deterministic.
Under these assumptions the baseline is
\begin{equation} \text{pass}@k \;=\; 1 - \big(1-p(x)\big)^{k} \label{eq:naive}
\end{equation}

\paragraph{Concept-conditioned sampling.} We instead condition generation on concepts.
Given a set of $M$ concepts $c=(c_1,\dots,c_M)$ sampled from $\pi_\theta(\cdot\,|\,x)$, define $p(x,c_j) = \mathbb{E}_{y\sim q(\cdot|x,c_j)}[\mathbbm{1}\{y\text{ correct}\}]$ and allocate $n = k/M$ rollouts to each concept (for the analysis we assume $M \mid k$, so that $n$ is an integer, the implementation splits the allocation as evenly as possible, \cref{sec:training_concept_generators}).
We additionally assume (iii) rollouts are independent \emph{across} concepts given $c$.
Note that the concepts themselves are \emph{not} independent as they are produced by a single autoregressive trajectory of $\pi_\theta$, but this does not affect the product below, which is taken over rollout outcomes conditionally on $c$.
The probability of obtaining at least one correct trajectory is then
\begin{equation} \text{pass}^{c}@k \;=\; 1 - \prod_{j=1}^{M}\big(1-p(x,c_j)\big)^{n} \;=\; 1 - \big(1-p_{\text{eff}}(x,c)\big)^{k},
\label{eq:passc}
\end{equation}
where we define the \emph{effective solve rate}
\begin{equation} p_{\text{eff}}(x,c) \;=\; 1 - \Big[\textstyle\prod_{j=1}^{M}\big(1-p(x,c_j)\big)\Big]^{1/M}.
\label{eq:peff}
\end{equation}
Equation~\eqref{eq:passc} puts concept-conditioned sampling on exactly the same footing as~\eqref{eq:naive}: $1-p_{\text{eff}}$ is the \emph{geometric mean of the per-concept failure probabilities}.
Moreover, $p_{\text{eff}}$ does not depend on $k$, so the comparison holds for any $k$.

\paragraph{When does conditioning help?} Define the yield of a concept as $\lambda(x,c_j) = -\log\big(1-p(x,c_j)\big)$, and let $\lambda_0 = -\log\big(1-p(x)\big)$ denote the baseline yield. Since $-\log\big(1-p_{\text{eff}}(x,c)\big) = \frac{1}{M}\sum_{j=1}^{M}\lambda(x,c_j) =: \bar\lambda(x,c)$, equations~\eqref{eq:naive} and~\eqref{eq:passc} read $\text{pass}@k = 1-e^{-k\lambda_0}$ and $\text{pass}^{c}@k = 1-e^{-k\bar\lambda(x,c)}$, so that
\begin{equation} \text{pass}^{c}@k \;>\; \text{pass}@k \qquad\Longleftrightarrow\qquad \bar\lambda(x,c) \;>\; \lambda_0 . \label{eq:criterion}
\end{equation}
Conditioning helps precisely when it raises the average per-concept yield above the baseline. This also means that every concept must earn its share of the allocation. Adding a further concept $c_{M+1}$ replaces $\bar\lambda$ by $\big(M\bar\lambda + \lambda(x,c_{M+1})\big)/(M+1)$, which improves $\text{pass}^{c}@k$ if and only if
\begin{equation} \lambda(x,c_{M+1}) \;>\; \bar\lambda(x,c). \label{eq:marginal}
\end{equation}
A concept that is merely below the current average is not neutral but actively harmful: it consumes $k/M$ rollouts while contributing less yield. This dilution effect is the central drawback of concept-conditioned sampling, and it is what any useful reward must penalise. We note in passing that, because $\lambda$ is convex in $p$, Jensen's inequality gives $p_{\text{eff}}(x,c) \ge \bar p(x,c) = \frac{1}{M}\sum_j p(x,c_j)$, with equality iff all $p(x,c_j)$ coincide, so $\bar p(x,c) > p(x)$ is sufficient for improvement, though \eqref{eq:criterion} is the exact criterion.

\paragraph{Why we evaluate on problems where the baseline is weak.}
The headroom available to any method is $1-\text{pass}@k = (1-p(x))^{k}$, which decays exponentially in $p(x)$. At $p(x)=0.1$ and $k=128$ the baseline already reaches $\text{pass}@k \approx 1-1.4\!\times\!10^{-6}$, leaving essentially nothing to win.
We therefore focus on problems for which $p(x)$ is small, which is exactly what our evaluation filter selects. Beyond this ceiling effect, there is a second reason to expect no gain when $p(x)$ is already high: conditioning on a concept restricts $q$ to a narrower region of the solution space, so when the unconditional model already covers that space well, steering it toward particular exploration directions would offer little upside.

\paragraph{Rollouts allocation.} Our uniform allocation $n_j = k/M$ is not optimal. Maximising $1-\prod_j(1-p(x,c_j))^{n_j}$ subject to $\sum_j n_j = k$ is equivalent to maximising the linear objective $\sum_j n_j \lambda(x,c_j)$, whose solution is degenerate: an oracle with knowledge of the $p(x,c_j)$ would assign the entire allocation to the single best concept, achieving $1-(1-p_{\max}(x,c))^{k}$ with $p_{\max}(x,c) = \max_j p(x,c_j)$.
This gives the chain
\begin{equation} \bar p(x,c) \;\le\; p_{\text{eff}}(x,c) \;\le\; p_{\max}(x,c), \label{eq:chain}
\end{equation}
whose three terms correspond respectively to mean concept quality, to the effective quality realised under uniform allocation, and to the quality realised under oracle allocation. Closing the second gap is naturally posed as a \emph{bandit problem} over concepts. We retain uniform allocation for throughput reasons, and leave adaptive allocation which would give better performance to future work.

\paragraph{Reward design.} The \emph{max-of-max} reward is unbiased for the objective we actually deploy: conditioned on the generated concepts $c$, we have $r_{\text{max-max}}\,|\,c \sim \text{Bernoulli}\big(\text{pass}^{c}@k\big)$, and in particular $\mathbb{E}\big[r_{\text{max-max}}\,|\,c\big] = \text{pass}^{c}@k$.
It is, however, a one-bit measurement: it collapses all $k$ rollout outcomes into a single indicator, leaving almost no basis for comparing concept sets within a group. In particular, whenever every set in a group succeeds or every set fails, the group-normalised advantages are identically zero and the update contributes no gradient which corresponds to the vanishing advantage regime documented for binary verifiable rewards~\citep{yu2026dapo}.

We therefore also consider the \emph{max-of-mean} reward, which averages the $k/M$ rollouts within each concept before aggregating across concepts, retaining $O(\log(k/M))$ bits rather than one. It rewards generating at least one concept with a high solve rate $p(x,c_j)$, and is a monotone estimate of $p_{\max}(x,c)$.
By~\eqref{eq:chain}, this means $r_{\text{max-mean}}$ is order preserving for the $\text{pass}^c@k$ attainable under \emph{oracle} allocation rather than under the uniform allocation we deploy: its bias is precisely the allocation gap $p_{\max}-p_{\text{eff}}$, and it would become the aligned reward under the adaptive allocation discussed above.
Two consequences of this mismatch are worth noting: by replacing the noisy-OR over concepts with a hard maximum, $r_{\text{max-mean}}$ does not credit sets that succeed through the combination of several moderate concepts and it inherits the upward bias of a maximum over noisy estimates ($\mathbb{E}\left[r_{\text{max-mean}}\,|\,x,c\right] \ge p_{\max}(x,c)$). We nonetheless adopt it because in our regime the resulting gain in reward informativeness outweighs this bias, as we show in \cref{sec:experiments}.

\paragraph{Choice of $M$.} The number of concepts is bounded by two competing effects. Additional concepts widen the search over exploration directions, but by~\eqref{eq:marginal} a concept improves $\text{pass}^{c}@k$ only if its yield exceeds the current average: a condition that marginal concepts increasingly fail as $M$ grows, so that dilution eventually dominates. At the same time, larger $M$ reduces the rollout count $n=k/M$ per concept, making each $\hat p(x,c_j)$ noisier and strengthening the upward bias of $r_{\text{max-mean}}$.

\paragraph{Estimating $\text{pass}^{c}@k$ from a larger rollout pool.}
In evaluation we generate $B$ rollouts and use the full pool to estimate performance at smaller allocations $k\leq B$, rather than discard $B-k$ outcomes and evaluate a single subset. As in \citet{chen2021evaluating}, averaging over all admissible $k$-subsets yields a lower-variance estimate of the corresponding pass probability. For clarity, suppose that $M$ divides both $B$ and $k$, let $b=B/M$ and $n=k/M$, and let $S_j$ be the number of successes among the $b$ rollouts assigned to concept $c_j$. An exactly unbiased estimator for the fixed balanced allocation is
\begin{equation}
\widehat{\text{pass}^{c}@k}_{\mathrm{strat}}
=1-\prod_{j=1}^{M}\frac{\binom{b-S_j}{n}}{\binom{b}{n}}.
\label{eq:stratified_passk}
\end{equation}
Indeed, $\frac{\binom{b-S_j}{n}}{\binom{b}{n}}$ is an unbiased estimator, based on $b$ observed rollouts, of the probability that $n$ rollouts conditioned on concept $c_j$ all fail.
Since answer rollouts are independent conditional on $c_1,\dots,c_M$, multiplying these per-concept failure probability estimators and subtracting the result from one, gives the estimator in \eqref{eq:stratified_passk}.

In our results we instead use the standard pooled estimator of \citet{chen2021evaluating}, $1-\binom{B-S}{k}/\binom{B}{k}$, where $S=\sum_j S_j$, for simplicity and direct comparability with the naive baseline.
This estimator is not exactly unbiased for a fixed balanced allocation when the $p(x,c_j)$ differ: it is unbiased for selecting a uniformly random $k$-subset of the pooled $B$ rollouts, which induces random per-concept counts $A=(A_1,\ldots,A_M)$ with a multivariate hypergeometric distribution and $\mathbb{E}[A_j]=n$. Writing $\lambda_j=\lambda(x,c_j)=-\log(1-p(x,c_j))$, Jensen's inequality gives
\begin{equation}
\mathbb{E}\!\left[\widehat{\text{pass@}k}_{\mathrm{Chen}}\mid c\right]
=1-\mathbb{E}_{A}\!\left[e^{-\sum_j A_j\lambda_j}\right]
\leq 1-e^{-n\sum_j\lambda_j}
=\text{pass}^{c}@k,
\label{eq:chen_conservative}
\end{equation}
so the pooled estimator is conservative in the uniform allocation setting. Moreover, if $v_\lambda=M^{-1}\sum_j(\lambda_j-\bar\lambda)^2$, using a second order approximation gives the following leading discrepancy:
\begin{equation}
\text{pass}^{c}@k-\mathbb{E}\!\left[\widehat{\text{pass@}k}_{\mathrm{Chen}}\mid c\right]
\simeq \frac{1}{2}\big(1-\text{pass}^{c}@k\big)\frac{k(B-k)}{B-1}v_\lambda,
\label{eq:chen_gap}
\end{equation}
which is second order in the variation of the per-concept yields. Their expectations coincide when all concepts have the same success probability, and the estimators themselves coincide exactly at $k=B$. Hence our headline pass@$128$ values with $B=128$ are unaffected. Moreover, in our low success regime the per-concept solve rates are small (\cref{sec:pos_biais}), so the correction in~\eqref{eq:chen_gap} is correspondingly small. We therefore retain the pooled Chen estimator as a simple, common estimator for all methods and allocations.

\section{Prompts}\label{app:prompts}
This appendix records the prompts used for concept generation, answer generation, baseline answer generation, and answer judging.

\subsection{Concept Generator Prompt}\label{app:cg_prompt}
The concept generator is prompted to first analyze the problem and then enumerate a short list of high-signal, non-duplicated concepts, theorems, and creative problem-solving techniques, while being explicitly instructed not to solve the problem or reveal the final answer. The system and user messages are shown below.
\begin{promptbox}[Concept Generator Prompt]
\begin{lstlisting}[basicstyle=\ttfamily,breaklines=true,breakatwhitespace=true,breakindent=0pt,breakautoindent=false,columns=fullflexible]
System:
You are an expert mathematician.
You will be given a mathematical problem. First, carefully analyze the problem:
identify what type of problem it is, what mathematical structures are present, and what approaches might work.
Then, list ALL relevant mathematical theorems, fundamental concepts, hints, and creative problem-solving techniques that could be useful for solving the problem.
Do NOT solve the problem. Do NOT compute the final answer.

OUTPUT FORMAT (MUST FOLLOW EXACTLY):
First, write your analysis of the problem.
Then, list concepts. Each concept line MUST start with exactly: ####
After '#### ' write ONE concept/theorem or a sentence for the creative technique, for example: #### your concept or technique here.
Each concept must be high-signal and specific enough (avoid generic items like 'algebra' or 'logical reasoning').
List between 1 and 10 concepts.
NO DUPLICATES OR NEAR DUPLICATES. The techniques or concepts should be new and different than previously generated ones.
Don't write anything after the end of the last concept.

User:
QUESTION:
{problem}
\end{lstlisting}
\end{promptbox}

\subsection{Concept-Conditioned Answer Generator Prompt}\label{app:ag_prompt}
The answer generator is prompted to produce a step-by-step solution to the problem while conditioning on a single concept, which is injected as a hint in the user message. The system and user messages are shown below.
\begin{promptbox}[Concept-Conditioned Answer Generator Prompt]
\begin{lstlisting}[basicstyle=\ttfamily,breaklines=true,breakatwhitespace=true,breakindent=0pt,breakautoindent=false,columns=fullflexible]
System:
You are an expert mathematician. Your task is to answer mathematical questions
with step-by-step solution.
Follow these instructions precisely:
Present your solution as a step-by-step process.
Explain each step clearly and concisely.
Use correct mathematical notation throughout your solution.
Output the final answer within \boxed{}.

User:
QUESTION:
{problem}

Here is a potentially useful hint: {concept}
Let's think step by step and output the final answer within \boxed{}.
\end{lstlisting}
\end{promptbox}

\subsection{Repeated-Sampling Baseline Prompt}
The repeated-sampling baseline uses the same answer-generator prompt as \cref{app:ag_prompt} but without any concept hint, so that all rollouts are drawn from the unconditioned answer generator. The system and user messages are shown below.
\begin{promptbox}[Repeated-Sampling Baseline Prompt]
\begin{lstlisting}[basicstyle=\ttfamily,breaklines=true,breakatwhitespace=true,breakindent=0pt,breakautoindent=false,columns=fullflexible]
System:
You are an expert mathematician. Your task is to answer mathematical questions with step-by-step solution.
Follow these instructions precisely:
Present your solution as a step-by-step process.
Explain each step clearly and concisely.
Use correct mathematical notation throughout your solution.
Output the final answer within \boxed{}.

User:
QUESTION:
{problem}

Let's think step by step and output the final answer within \boxed{}.
\end{lstlisting}
\end{promptbox}

\subsection{Generic Prompt Baseline}\label{app:generic_prompt}
The generic prompt-modification baseline uses the exact same concept-conditioned answer-generator prompt (\cref{app:ag_prompt}), but replaces the problem-specific concept in the \verb|{concept}| slot with one of ten fixed, problem-agnostic hints. The ten generic hints are:
\begin{promptbox}[Generic Hints]
\begin{lstlisting}[basicstyle=\ttfamily,breaklines=true,breakatwhitespace=true,breakindent=0pt,breakautoindent=false,columns=fullflexible]
Try a different angle than the obvious one.
Use the classical, textbook approach to solve this.
Think outside the box and look for a creative shortcut.
Break the problem into smaller, simpler subproblems.
Look for symmetry, patterns, or invariants you can exploit.
Work backwards from the desired result to the given information.
Consider small or special cases first to build intuition.
Translate the problem into algebra and manipulate the equations carefully.
Reformulate the problem geometrically or visually.
Estimate or bound the answer before computing it exactly.
\end{lstlisting}
\end{promptbox}

\subsection{Judge Prompt}\label{app:judge_prompt}
We grade answers with an LLM judge that compares the student answer against the ground-truth answer and returns a binary correctness verdict. The system and user messages are shown below.
\begin{promptbox}[Judge Prompt]
\begin{lstlisting}[basicstyle=\ttfamily,breaklines=true,breakatwhitespace=true,breakindent=0pt,breakautoindent=false,columns=fullflexible]
System:
You are a math verifier. Compare between the student answer and the given correct ground truth.
Reply ONLY 'correct' or 'incorrect'.

User:
problem: {problem}

student answer: {answer}

ground truth: {ground_truth}

verdict:
\end{lstlisting}
\end{promptbox}

\subsection{Leakage-Detection Judge Prompt}\label{app:leakage_prompt}
To audit for answer leakage, we use a separate LLM judge that, given the ground-truth answer and a single concept, decides whether the concept states or reveals the final answer. The system and user messages are shown below.
\begin{promptbox}[Leakage-Detection Judge Prompt]
\begin{lstlisting}[basicstyle=\ttfamily,breaklines=true,breakatwhitespace=true,breakindent=0pt,breakautoindent=false,columns=fullflexible]
System:
You are a strict grader whose job is to detect answer leakage.
You are given the GROUND-TRUTH final answer to a math problem and a single
CONCEPT. A concept is a hint or exploration strategy that is meant to guide someone toward solving the problem WITHOUT revealing the final answer.
Decide whether the concept reveals the answer. Answer 'yes' if the concept explicitly states the ground-truth answer, or states a numeric value or expression that is mathematically equivalent to the ground-truth answer.
Answer 'no' if the concept only describes methods, strategies, intuitions, reformulations, or partial setups that do not state the final answer value.
Do not try to solve the problem yourself; judge only whether the answer value is present in the concept.
Reply ONLY 'yes' or 'no'.

User:
ground truth answer: {ground_truth}

concept: {concept}

Does the concept state or reveal the ground-truth answer (or a value
mathematically equivalent to it)? Reply ONLY 'yes' or 'no'.
verdict:
\end{lstlisting}
\end{promptbox}

\section{Training Details}\label{app:training_details}
This appendix gives the full training and generation configuration. The main text summarizes the setup in \cref{sec:experiments}. We implement our training loop on top of verl
\citep{sheng2024hybridflow}.

\begin{table}[!ht]
\centering
\caption{Training and generation hyperparameters.}
\label{tab:training_hyperparams}
\begin{tabular}{lll}
\toprule
\textbf{Component} & \textbf{Hyperparameter} & \textbf{Value} \\
\midrule
CG training & Learning rate & $1\times 10^{-6}$ \\
CG training & Warmup steps & $50$ \\
CG training & Optimizer & AdamW \\
CG training & AdamW betas & $(0.9, 0.99)$ \\
CG training & KL coefficient & $0.001$ \\
CG training & PPO clip low / high & $0.2$ / $0.272$ \\
CG training & Train batch size & $64$ \\
CG training & PPO mini-batch size & $32$ \\
CG training & Loss aggregation & token-mean \\
CG generation & Temperature & $0.8$ \\
CG generation & top-$p$ / top-$k$ & $0.95$ / $100$ \\
CG generation & Max prompt / response length & $2048$ / $2048$ \\
RL rollout & Trajectories per problem $G$ & $8$ \\
RL rollout & AG rollout allocation per trajectory $B$ & $128$ \\
AG generation & Temperature & $1.0$ \\
AG generation & top-$p$ / top-$k$ & $0.95$ / $-1$ \\
AG generation & Max prompt / response length & $2048$ / $4096$ \\
Judge & Model & Qwen2.5-32B-Instruct \\
Judge & Decoding & Greedy \\
\bottomrule
\end{tabular}
\end{table}

\subsection{Training Objective}
The main text (\cref{sec:training_concept_generators}) gives the group-relative advantage $A_i = r_i - \frac{1}{G}\sum_{j=1}^{G} r_j$ used to train the concept generator $\pi_\theta$. Here we state the complete GRPO-style objective, as used in Olmo3 \citep{olmo2025olmo}, including the clipped policy-gradient term and the KL penalty to a fixed reference policy:

\begin{equation*}
\resizebox{0.99\textwidth}{!}{$
\mathcal{J}(\theta)
= \frac{1}{\sum_{i=1}^{G} |y_i|}
  \sum_{i=1}^{G} \sum_{t=1}^{|y_i|}
  \left[
  \min\!\left(
  \frac{\pi_{\theta_{\mathrm{old}}}\!\left(y_{i,t}\mid x, y_{i,<t}\right)}{\pi_{\theta_{\mathrm{old}}}^{\mathrm{vllm}}\!\left(y_{i,t}\mid x, y_{i,<t}\right)}, \rho\right)
  \min\!\left(
  w_{i,t}\, A_{i,t},\;
  \operatorname{clip}\!\left(
  w_{i,t},\;
  1 - \varepsilon_{\mathrm{low}},\;
  1 + \varepsilon_{\mathrm{high}}
  \right) A_{i,t}
  \right)
  - \beta\, \mathbb{D}_{\mathrm{KL}}\!\left(\pi_\theta \,\|\, \pi_{\mathrm{ref}}\right)
  \right]
$}
\end{equation*}

Here $G$ is the group size, i.e.\ the number of concept trajectories sampled per problem, and $y_i$ is the token sequence of trajectory $i$ (with length $|y_i|$), so the objective is normalized per token (token-mean) across the group. The ratio $w_{i,t}=\pi_\theta(y_{i,t}\mid x,y_{i,<t})/\pi_{\theta_{\mathrm{old}}}(y_{i,t}\mid x,y_{i,<t})$ is the usual token-level policy ratio, and the leading $\min(\cdot,\rho)$ factor is a truncated importance-sampling correction between the rollout policy $\pi_{\theta_{\mathrm{old}}}^{\mathrm{vllm}}$ that generated the trajectory under vLLM and the same policy $\pi_{\theta_{\mathrm{old}}}$ as recomputed by the training backend, capped at $\rho=2.0$. We use an asymmetric (clip-higher) clipping range $\varepsilon_{\mathrm{low}}=0.2$, $\varepsilon_{\mathrm{high}}=0.272$, and do not normalize advantages by their standard deviation. The advantage $A_{i,t}=A_i$ is the trajectory-level group-relative advantage broadcast to every token of trajectory $i$, where the trajectory reward $r_i$ is the aggregate scalar obtained from the binary ($0/1$) correctness scores of the answer-generator rollouts for that concept trajectory (max-of-max or max-of-mean, \cref{sec:training_concept_generators}). The KL penalty $\mathbb{D}_{\mathrm{KL}}(\pi_\theta \,\|\, \pi_{\mathrm{ref}})$ uses the low-variance estimator $\frac{\pi_{\mathrm{ref}}}{\pi_\theta} - \log\frac{\pi_{\mathrm{ref}}}{\pi_\theta} - 1$ with coefficient $\beta=0.001$; the reference policy $\pi_{\mathrm{ref}}$ is the concept generator before any RL training and is held fixed throughout.

\subsection{Dataset Filtering}
We filter DeepMath-103k \citep{he2026deepmath} using the answer generator (Qwen2.5-32B-Instruct) and retain training problems where the AG obtains less than $5\%$ success rate estimated with 128 rollouts, i.e.\ at most 6 correct answers out of 128. Success is judged by Qwen2.5-32B-Instruct, and the filtering rollouts use the same AG decoding parameters as in \cref{tab:training_hyperparams} (temperature $1.0$, top-$p=0.95$, top-$k=-1$). From the filtered pool we reserve a held-out evaluation set of 1k problems with $0\%$ success rate estimated with 128 rollouts, disjoint from training; this leaves $10{,}776$ problems for training. We filter at $5\%$ success rate for training, while requiring a $0\%$ success rate for evaluation, ensuring that the training set still provides a learnable signal.

To evaluate out-of-distribution generalization, we filter Omni-MATH-2 \citep{ballon2026benchmarkssaturatewhen}, itself a filtered version of Omni-MATH \citep{gao2025omni}, with the same procedure, keeping only problems with $0\%$ success rate estimated with 128 rollouts. This yields $1{,}382$ problems, which are used only for evaluation.

\subsection{Concept Parsing}
The CG prompt requires every concept to start with the delimiter \verb|####|. Parsing is regex-based and lenient: any analysis or preamble before the first \verb|####| is ignored, whitespace-only spans are dropped, and malformed text simply yields no matches rather than raising an error. We keep at most the first ten parsed concepts per trajectory ($\texttt{max\_concepts\_per\_trajectory}=10$); when a trajectory exceeds this, its response tokens are truncated on-policy at the cut point so that the retained tokens keep their original conditional probabilities.

\textbf{Empty concept lists.} A trajectory with no valid \verb|####| marker parses to an empty list. During training such trajectories generate no answer rollouts and receive a fixed penalty reward of $-0.1$, which mildly discourages degenerate outputs; during validation they are replaced by a single empty concept so the pipeline still runs. We track the fraction of empty trajectories as a metric; in practice it stays at roughly $0\%$ throughout training, so parsing works out of the box.

\textbf{Duplicates.} Duplicate concepts are not removed for training or answer generation: identical concepts, if they appear, are treated as distinct list positions, i.e. we don't apply any deduplication.

\textbf{Length limits.} If a single concept would exceed the remaining prompt allowed length (maximum prompt length minus the problem and template, with a small safety margin), it is truncated to fit.

\section{Additional Results}\label{app:additional_results}
This appendix presents additional empirical results omitted from the main paper for space, including sanity checks, a direct-training comparison, full pass@k curves with confidence intervals, and analyses of concept position and count.

\subsection{Training the Smaller Model Directly}\label{app:direct_training}
\begin{wrapfigure}{r}{0.42\linewidth}
  \centering
  \includegraphics[width=\linewidth]{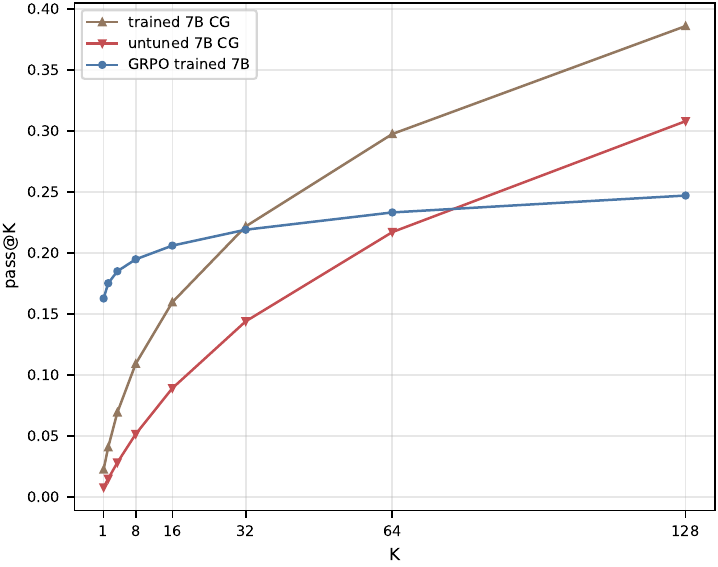}
  \caption{Held-out pass@$k$ on the 1k hard set: a 7B trained to answer directly, our trained 7B CG steering the 32B, and the untuned 7B CG steering the 32B.}
  \label{fig:ablation1}
\end{wrapfigure}
A second question is whether the CG+AG setup is even necessary: could we instead simply train the small model to answer the hard problems on its own and skip the answer generator altogether? This matters because in our setting the AG is frozen and possibly closed-source, so the only trainable alternative is the small model itself.
To test this, we train an identical 7B model with the same RL algorithm, training split, and LLM judge, but with a direct answer objective, and compare it against our CG-steered 32B across the rollout allocation (\cref{fig:ablation1}).
Moreover, even the \emph{untuned} 7B CG steering the 32B reaches $28.9\%$ pass@128, already above the trained direct-answer model. \textbf{Training the small model to answer directly does not recover the performance of using it as a concept generator for a larger model: the value of our approach is injecting exploration into a stronger frozen answer generator at scale, not having the small model solve problems by itself.}
The direct-answer model does reach $16.3\%$ pass@1 (and $24.7\%$ pass@128) on problems where the 32B AG scored $0\%$ at 128 rollouts; this reflects the finite-sample nature of the filter (a different model recovers a fraction of these problems, \cref{app:filtering_variance}) rather than genuine $0\%$ difficulty.

\subsection{Position Bias Under Trajectory-Level Rewards}\label{sec:pos_biais}
\begin{figure}[htbp]
  \centering
  \includegraphics[width=0.42\linewidth]{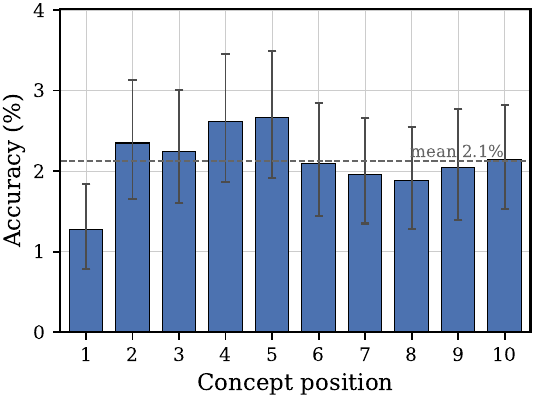}
  \caption{Per-concept accuracy by position in the trajectory (trained 7B CG, max-of-mean). Error bars are $95\%$ bootstrap CIs over the held-out problems ($10{,}000$ resamples, $2.5/97.5$ percentiles).}
  \label{fig:position_bias}
\end{figure}
Because the reward is assigned to the whole concept trajectory rather than to individual concepts (\cref{sec:training_concept_generators}), weak concepts could in principle free-ride on strong ones, and the model could learn to pad later positions with filler. We check for this by measuring, for each position in the trajectory, the mean accuracy of the AG rollouts conditioned on the concept at that position (the same per-concept mean used by max-of-mean), averaged over the trajectories that contain a concept at that position. Coverage stays high throughout: every position is present in at least $\sim\!89\%$ of trajectories (positions receive between $11{,}580$ and $13{,}084$ rollouts), so the pattern is not an artifact of missing late concepts.
Per-position accuracy varies only within a narrow band: it ranges from $1.3\%$ at the first concept to $2.7\%$ at the fifth, with all remaining positions clustered near $2\%$. Crucially, no position is dead: every position yields correct solutions and none dominates.
The best position is only about twice the weakest, while positions~2--10 have broadly comparable estimated accuracies and overlapping bootstrap confidence intervals.
The \emph{first} concept is the weakest, suggesting that useful ideas are not concentrated at the start.
This supports our use of trajectory-level credit: it does not degenerate into a few useful positions plus dead ones.

\subsection{Answer Leakage}\label{app:answer_leakage}
A natural worry is reward hacking: the concept generator could learn to solve each problem then pass the final answer into its ``concepts,'' in which case the downstream gains would reflect answer leakage rather than better exploration.
We test this directly by checking whether the ground-truth answer is present in the generated concepts. For each held-out problem we run an LLM judge (Qwen2.5-32B-Instruct) over each parsed concept and ask whether it states, or gives a value mathematically equivalent to, the ground-truth answer; the judge prompt is given in \cref{app:leakage_prompt}. We deliberately avoid exact string matching, which produces spurious hits when the ground-truth answer is a short value that happens to coincide with unrelated text in a concept.
We find that only $0.44\%$ of concepts are flagged as containing the answer, and just $3\%$ of problems have at least one flagged concept. This confirms that the CG steers exploration rather than leaking solutions.

\subsection{Effect of the Number of Concepts}\label{app:num_concepts}
\begin{wrapfigure}[20]{r}{0.42\linewidth}
  \centering
  \includegraphics[width=\linewidth]{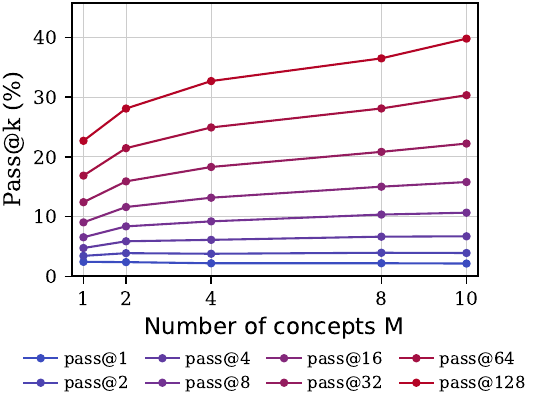}
  \caption{Pass@$k$ vs.\ number of concepts $M$ at a fixed total allocation ($B=128$ split across the $M$ concepts); concept sets are nested as $M$ increases, and $k$ ranges from $1$ to $128$ (blue to red). Trained 7B CG (max-of-mean) steering the 32B AG over the 1k held-out set.}
  \label{fig:num_concepts}
\end{wrapfigure}
After training, the CG almost always still emits close to ten concepts (we truncate to the first ten), so the count does not collapse during training. To separate the value of using \emph{multiple} concepts from the value of simply spending more compute, we hold the total answer allocation fixed. For each held-out problem, we first sample one generated concept uniformly at random, then incrementally sample additional concepts without replacement to form nested sets of sizes $M\in\{1,2,4,8,10\}$. Thus every larger set retains all concepts from the smaller settings. At each $M$, we split the same $B=128$ rollout allocation as evenly as possible across the selected concepts (so $M=1$ concentrates all $128$ rollouts on one concept, whereas $M=10$ assigns $12$ or $13$ rollouts to each).
Under this nested comparison, more concepts monotonically improve performance for $k\geq4$: pass@$128$ rises from $22.70\%$ with one concept to $28.10\%$, $32.70\%$, $36.50\%$, and $39.80\%$ with $2$, $4$, $8$, and $10$ concepts, respectively. pass@$64$ similarly rises from $16.87\%$ to $30.34\%$ (\cref{fig:num_concepts}).
In contrast, pass@$1$ remains in the narrow range $2.15$--$2.44\%$. This widening benefit with $k$ is the signature of improved \emph{exploration} rather than single-shot accuracy: spreading a fixed allocation over additional, diverse concepts covers more of the solution space than repeatedly sampling from one. The incremental gains diminish as $M$ grows but remain positive through ten concepts. Because total compute is fixed and the concept sets are nested, the trend cannot be explained by extra sampling or by independently redrawing a more favorable subset at each $M$: it directly supports our choice to generate many concepts rather than one.

\subsection{Iterative Guided-Sampling Protocol on the Hard Subsets}\label{app:guided_sampling_hard}
For a like-for-like comparison with our inference-time procedure (\cref{sec:concept_guided_sampling}), we run the original iterative guided-sampling protocol of \citet{handa2026guidedsampling} on the \emph{same} model-specific hard MATH subsets, using the \emph{same} favorable exploratory decoding parameters for the answer generator (temperature $1.0$, top-$p=0.95$, top-$k=-1$). Only the concept-generation procedure differs: concepts are produced one at a time with the original iterative protocol rather than in a single trajectory. \Cref{tab:guided_sampling_hard} reports the resulting pass@50 differences over the repeated-sampling baseline; the baseline pass@50 values are identical to those in \cref{tab:concepts_ours}.

The original protocol does yield some improvements on these hard subsets once the answer generator is allowed to explore, but they are consistently smaller than ours (\cref{tab:concepts_ours}): its largest gain is $+6.8$ points (vs.\ our $+9.7$), and several CG/AG pairs still regress. We attribute the gap to concept count: because the iterative protocol samples concepts one by one, it produces at most $1.5$ concepts per problem on average (for the $32$B generator), whereas our single trajectory procedure produces up to~$\sim\!10$.

\begin{table}[!ht]
\centering
\caption{\label{tab:guided_sampling_hard}Iterative guided-sampling protocol~\citep{handa2026guidedsampling} (without training) on the same model specific hard MATH subsets (Qwen2.5-Instruct) as \cref{tab:concepts_ours}. We report pass@50 concept-guided differences across CG/AG size pairs.}
\begin{tabular}{lccccc}
\toprule
& \multicolumn{5}{c}{AG} \\
\cmidrule(lr){2-6}
& 1.5B & 3B & 7B & 14B & 32B \\
\midrule
Baseline pass@50 & 18.4 & 16.6 & 11.0 & 12.6 & 11.8 \\
\midrule
\addlinespace
\textbf{CG / AG} &  \multicolumn{5}{l}{\textbf{Difference (Concept - Baseline) pass@50}} \\
\addlinespace
\quad 1.5B & +0.8 & +1.2 & +4.1 & +0.5 & +1.9 \\
\quad 3B & +0.7 & +0.7 & +5.3 & +1.0 & +3.1 \\
\quad 7B & -0.1 & +2.5 & +4.2 & -1.0 & +4.3 \\
\quad 14B & +1.1 & -0.3 & +5.4 & +2.3 & +2.0 \\
\quad 32B & +0.2 & -1.7 & +6.8 & +1.0 & +4.7 \\
\bottomrule
\end{tabular}
\end{table}

\subsection{Why Naive Resampling Is Nonzero on the ``\texorpdfstring{$0\%$}{"0\%"}'' Held-Out Set}
\label{app:filtering_variance}
We build the held-out set by drawing $B=128$ answers per problem with the answer generator and keeping the problems with zero correct samples. This is a finite-sample estimate of the AG success probability $p$, not a guarantee that $p=0$. A problem with small but nonzero $p$ passes the filter with probability $(1-p)^{128}$, yet an independent fresh draw of $128$ answers solves it at least once, with probability $1-(1-p)^{128}$.

Empirically, the naive baseline reaches pass@128 ($\approx19\%$) on this filtered set (\cref{tab:main_results}). Because this estimate is based on only (128) fresh evaluation rollouts per problem, it may be sensitive to the particular sampling draw. We therefore re-estimate it below using a much larger rollout allocation.

\textbf{Re-estimating the naive pass@128.}
The main-table pass@128 is computed with the unbiased estimator~\citep{chen2021evaluating} on per-problem success counts $s_i$ out of $B=128$ rollouts. At $k=B=128$ this estimator degenerates to the plain solve rate $\mathbbm{1}[s_i\ge1]$, so it carries the full Bernoulli seed noise and could in principle over- or under-shoot the true value. To estimate this variance and make sure our results are well beyond the seed variance of the naive sampling baseline, we draw a fresh batch of $1280$ answers per problem, generated as $10$ independent seeds of $128$ answers each (and independent of the filtering seed), and re-estimate the naive pass@128 with three estimators (\cref{tab:naive_pass128_variance}): (i) the unbiased estimator with $k=128<B=1280$, applied to all $1280$ pooled rollouts, which analytically averages over rollout noise (equivalent to averaging over all $\binom{1280}{128}$ subsets); (ii) a Monte Carlo estimate that draws $10{,}000$ random $128$-rollout subsets and reports the mean and standard deviation of the plain solve rate; and (iii) a direct seed-level estimate that reports the mean and standard deviation across the $10$ independent seeds.

The unbiased estimator gives $19.46\%$, and the Monte Carlo estimate agrees closely at $19.47\%$, with a standard deviation of $0.94$ points across random $128$-rollout subsets. The mean across the $10$ independent seeds is likewise $19.44\%$, with a standard deviation of $1.43$ points across seeds. Naive repeated sampling therefore recovers $\approx 19.5\%$ of the ``$0\%$'' held-out problems on a fresh seed.

\begin{table}[!ht]
\centering
\caption{\label{tab:naive_pass128_variance}Re-estimated naive pass@128 on the ``$0\%$'' held-out set (fresh batch of $B=1280$ rollouts per problem).}
\begin{tabular}{lcc}
\toprule
Estimator & pass@128 (\%) & std (\%) \\
\midrule
Unbiased (all $1280$ rollouts, $k=128$)        & $19.46$ & --- \\
Monte Carlo ($10{,}000$ random $128$-subsets)  & $19.47$ & $0.94$ \\
Independent seeds ($10\times128$ rollouts)        & $19.44$ & $1.43$ \\
\bottomrule
\end{tabular}
\end{table}

\subsection{Extended Pass@k Curves and Confidence Intervals}\label{app:extended_curves}
\begin{figure}[t]
  \centering
  \includegraphics[width=0.99\linewidth]{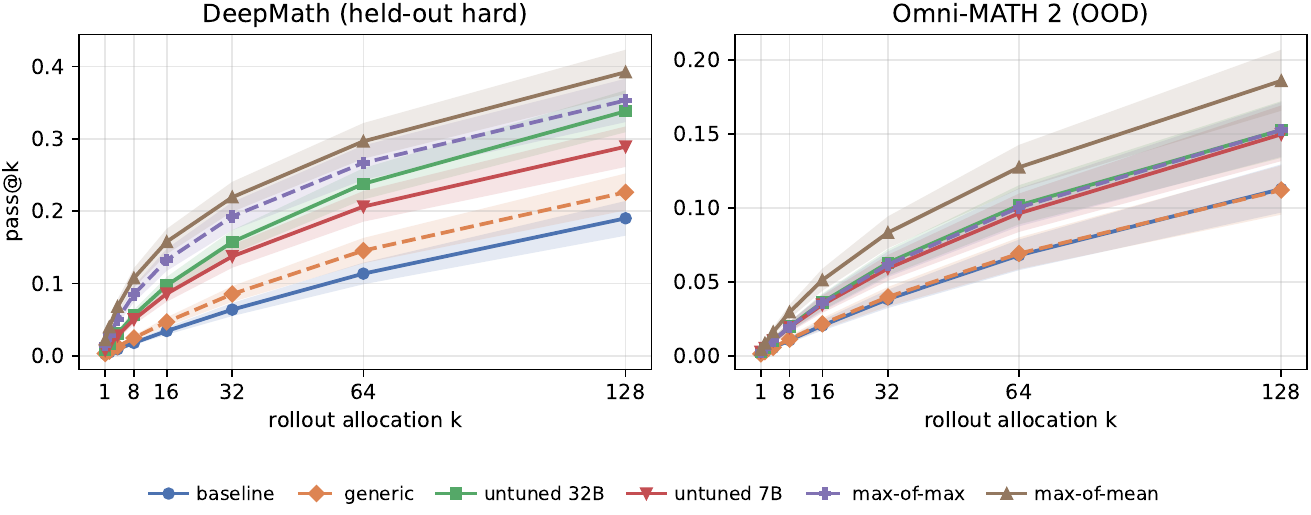}
  \caption{Pass@k vs.\ rollout allocation $k$ for all methods, one panel per dataset (DeepMath held-out hard, Omni-MATH~2), with shaded $95\%$ bootstrap confidence bands (bootstrap over held-out problems).
  }
  \label{fig:passk_curves}
\end{figure}

The main table (\cref{tab:main_results}) reports point estimates at $k\in\{1,64,128\}$. Here we report the full dependence on the rollout allocation together with estimation uncertainty, in two complementary forms: full pass@k curves with confidence bands (\cref{fig:passk_curves}), and per-method gains over the naive baseline with paired-bootstrap confidence intervals (\cref{tab:passk_ci}).

\textbf{Full curves with confidence bands.}
\Cref{fig:passk_curves} plots pass@k against the rollout allocation $k$ for every method, with one panel per dataset (DeepMath held-out hard, Omni-MATH~2 OOD) and a shaded $95\%$ bootstrap confidence band around each curve. This shows the entire dependence on $k$---superseding the three snapshot columns of the main table---together with estimation uncertainty, and makes the method ordering, and where bands overlap, visible at a glance.

\textbf{Gains over the naive baseline.}
Because all methods are scored on the same held-out problems, the quantity of interest is the per-problem gain over naive repeated sampling rather than the absolute pass@k. \Cref{tab:passk_ci} reports, for each method, the difference $\Delta\text{pass@}k$ relative to naive at $k\in\{1,64,128\}$ with a $95\%$ paired-bootstrap interval; an interval that excludes $0$ indicates a statistically significant improvement.
On DeepMath, every method improves significantly over naive at all allocations, and the gains grow sharply with $k$---from under two points at pass@1 to double digits at pass@128---the signature of better \emph{exploration} rather than single-shot accuracy. The ordering is consistent across allocations, with the trained max-of-mean policy strongest ($+20.2$ points at pass@128), followed by max-of-max ($+16.3$), the untuned 32B CG ($+14.8$), the untuned 7B CG ($+9.9$), and the generic prompt-modification baseline ($+3.6$).
On the Omni-MATH~2 OOD set the gains are smaller but the concept-guided methods remain significant at every allocation (max-of-mean $+7.3$ points at pass@128, intervals excluding $0$), whereas the generic prompt-modification baseline is statistically indistinguishable from naive throughout (e.g.\ $-0.07~[-1.88,+1.74]$ at pass@128). Problem-specific concepts, and especially the trained max-of-mean policy, thus yield significant gains that generic prompt perturbation does not, and these gains persist out of distribution.

\textbf{Estimation procedure.}
For each problem $i$ and each method, we generate $B=128$ answer rollouts. For concept-guided methods, we distribute these rollouts as evenly as possible across the $M$ concepts and pool the resulting candidates. Let $s_i$ denote the total number of correct answers in this pooled set. We then compute the standard per-problem estimate $\widehat{\text{pass@}k}_i = 1-\binom{B-s_i}{k}/\binom{B}{k}$ and report pass@k as its mean over the corresponding held-out set.

Confidence intervals come from a bootstrap over problems: we resample the $\sim$1000 problem indices with replacement, recompute the mean, repeat $10{,}000$ times, and take the $2.5/97.5$ percentiles. For the gain of one method over another we use a \emph{paired} bootstrap---resampling problems once and taking the difference of the two means on that same resample---since both methods are evaluated on identical problems. (At $k=B=128$ the estimator degenerates to the plain solve rate; see \cref{app:filtering_variance} for when extra rollouts, rather than seeds, are needed to stabilize it.)

\begin{table}[!ht]
\centering
\caption{\label{tab:passk_ci}Gain over naive repeated sampling, $\Delta\text{pass@}k$ (\%), with $95\%$ paired-bootstrap confidence intervals (over held-out problems). Cells are formatted as $\Delta$~$[\text{lo},\text{hi}]$; an interval excluding $0$ indicates a significant improvement.}
\resizebox{\textwidth}{!}{%
\begin{tabular}{lccc}
\toprule
Method (vs.\ naive) & $\Delta$pass@1 & $\Delta$pass@64 & $\Delta$pass@128 \\
\midrule
\multicolumn{4}{l}{\emph{DeepMath (held-out hard)}}\\
\quad Generic prompt modification & $+0.09$~$[+0.05, +0.13]$ & $+3.18$~$[+1.45, +4.87]$ & $+3.60$~$[+0.80, +6.30]$ \\
\quad Untuned CG (7B) & $+0.54$~$[+0.40, +0.68]$ & $+9.27$~$[+7.15, +11.32]$ & $+9.90$~$[+6.90, +12.80]$ \\
\quad Untuned CG (32B) & $+0.59$~$[+0.47, +0.71]$ & $+12.42$~$[+10.29, +14.54]$ & $+14.80$~$[+11.70, +17.90]$ \\
\quad max-of-max & $+1.25$~$[+1.02, +1.50]$ & $+15.31$~$[+12.93, +17.69]$ & $+16.30$~$[+13.00, +19.60]$ \\
\quad max-of-mean & $+1.95$~$[+1.59, +2.33]$ & $+18.29$~$[+15.91, +20.73]$ & $+20.20$~$[+17.00, +23.40]$ \\
\addlinespace
\multicolumn{4}{l}{\emph{Omni-MATH~2 (OOD)}}\\
\quad Generic prompt modification & $+0.01$~$[-0.02, +0.04]$ & $+0.12$~$[-0.99, +1.21]$ & $-0.07$~$[-1.88, +1.74]$ \\
\quad Untuned CG (7B) & $+0.14$~$[+0.09, +0.19]$ & $+2.83$~$[+1.63, +4.03]$ & $+3.69$~$[+1.74, +5.57]$ \\
\quad Untuned CG (32B) & $+0.13$~$[+0.09, +0.18]$ & $+3.37$~$[+2.14, +4.62]$ & $+3.98$~$[+2.03, +5.93]$ \\
\quad max-of-max & $+0.13$~$[+0.09, +0.18]$ & $+3.23$~$[+1.98, +4.45]$ & $+3.98$~$[+2.03, +5.93]$ \\
\quad max-of-mean & $+0.30$~$[+0.23, +0.39]$ & $+5.94$~$[+4.60, +7.35]$ & $+7.31$~$[+5.28, +9.41]$ \\
\bottomrule
\end{tabular}%
}
\end{table}

\textbf{Trained 7B CG vs.\ untuned 32B CG.}
A central claim of \cref{sec:experiments} is that the \emph{trained} 7B concept generator beats the \emph{untuned} 32B one, despite steering the same frozen answer generator with far fewer parameters. We test this gap directly with the same paired bootstrap over held-out problems, taking the per-problem difference of the two unbiased pass@k estimates on identical problems (\cref{tab:passk_ci_pairwise}). The trained max-of-mean CG significantly outperforms the untuned 32B CG at every allocation on both datasets (all $95\%$ intervals exclude $0$) by $+5.4~[+2.4,+8.6]$ points at pass@128 on DeepMath and $+3.3~[+1.2,+5.4]$ on Omni-MATH~2.

\begin{table}[!ht]
\centering
\caption{\label{tab:passk_ci_pairwise}Direct gain of the RL trained 7B CG over the untuned 32B CG, $\Delta\text{pass@}k$ (\%), with $95\%$ paired-bootstrap confidence intervals (over held-out problems). Cells are formatted as $\Delta$~$[\text{lo},\text{hi}]$; an interval excluding $0$ indicates a significant improvement.}
\resizebox{\textwidth}{!}{%
\begin{tabular}{lccc}
\toprule
Method (vs.\ untuned 32B CG) & $\Delta$pass@1 & $\Delta$pass@64 & $\Delta$pass@128 \\
\midrule
\multicolumn{4}{l}{\emph{DeepMath (held-out hard)}}\\
\quad max-of-max & $+0.67$~$[+0.46, +0.88]$ & $+2.88$~$[+0.67, +5.11]$ & $+1.50$~$[-1.60, +4.60]$ \\
\quad max-of-mean & $+1.37$~$[+1.05, +1.70]$ & $+5.87$~$[+3.70, +8.11]$ & $+5.40$~$[+2.40, +8.60]$ \\
\addlinespace
\multicolumn{4}{l}{\emph{Omni-MATH~2 (OOD)}}\\
\quad max-of-max & $+0.00$~$[-0.04, +0.05]$ & $-0.14$~$[-1.46, +1.15]$ & $+0.00$~$[-2.03, +2.03]$ \\
\quad max-of-mean & $+0.17$~$[+0.10, +0.26]$ & $+2.57$~$[+1.16, +4.00]$ & $+3.33$~$[+1.16, +5.43]$ \\
\bottomrule
\end{tabular}%
}
\end{table}

\textbf{Effect of RL training (trained vs.\ untuned 7B CG).}
Since the untuned and trained 7B CG are the \emph{same} model before and after RL, their difference isolates the effect of training itself. Under the same paired bootstrap (\cref{tab:passk_ci_vs7b}), training yields significant gains at every allocation on DeepMath (max-of-mean $+10.3~[+7.2,+13.4]$ points at pass@128; max-of-max $+6.4~[+3.4,+9.4]$). Out of distribution on Omni-MATH~2, max-of-mean stays significant at all allocations ($+3.6~[+1.5,+5.6]$ at pass@128), whereas max-of-max is indistinguishable from the untuned 7B CG (intervals include $0$). Training therefore improves concept quality beyond what better prompting and filtering achieve, and only the max-of-mean objective carries this improvement out of distribution.

\begin{table}[!ht]
\centering
\caption{\label{tab:passk_ci_vs7b}Gain of the RL-trained 7B CG over the untuned 7B CG (the same model before training), $\Delta\text{pass@}k$ (\%), with $95\%$ paired-bootstrap confidence intervals (over held-out problems). Cells are formatted as $\Delta$~$[\text{lo},\text{hi}]$; an interval excluding $0$ indicates a significant improvement.}
\resizebox{\textwidth}{!}{%
\begin{tabular}{lccc}
\toprule
Method (vs.\ untuned 7B CG) & $\Delta$pass@1 & $\Delta$pass@64 & $\Delta$pass@128 \\
\midrule
\multicolumn{4}{l}{\emph{DeepMath (held-out hard)}}\\
\quad max-of-max & $+0.72$~$[+0.48, +0.96]$ & $+6.04$~$[+3.85, +8.27]$ & $+6.40$~$[+3.40, +9.40]$ \\
\quad max-of-mean & $+1.42$~$[+1.08, +1.78]$ & $+9.03$~$[+6.77, +11.39]$ & $+10.30$~$[+7.20, +13.40]$ \\
\addlinespace
\multicolumn{4}{l}{\emph{Omni-MATH~2 (OOD)}}\\
\quad max-of-max & $0.00$~$[-0.06, +0.06]$ & $+0.40$~$[-0.90, +1.68]$ & $+0.29$~$[-1.74, +2.32]$ \\
\quad max-of-mean & $+0.17$~$[+0.09, +0.25]$ & $+3.11$~$[+1.74, +4.49]$ & $+3.62$~$[+1.52, +5.64]$ \\
\bottomrule
\end{tabular}%
}
\end{table}

\subsection{Diversity of the Generated Reasoning}\label{app:diversity}
A central claim of our approach is that concept guidance improves \emph{exploration}: it should yield genuinely different solution attempts rather than the near-duplicates produced by decoding noise alone. We test this directly by measuring the semantic diversity of the generated reasoning on the DeepMath val-hard held-out set.

\textbf{Setup.}
For each problem we collect the model's rollouts and embed only the chain-of-thought excluding the problem statement and the final answer, so that the score reflects the diversity of the \emph{reasoning} itself rather than shared prompt text or a common answer. We use Qwen3-Embedding-8B \citep{qwen3embedding} as the sentence embedder. On the resulting set of embeddings we compute the Vendi Score (VS)~\citep{friedman2023the} with an RBF kernel $k(x,y)=\exp\!\big(-\lVert x-y\rVert^2/(2\sigma^2)\big)$. The VS is the exponential of the Shannon entropy of the eigenvalues of the normalized kernel matrix and can be read as the \emph{effective number of distinct rollouts}, so a higher VS means more diverse reasoning. We compute VS per problem and average over the held-out set.

\begin{figure}[t]
  \centering
  \includegraphics[width=0.4\linewidth]{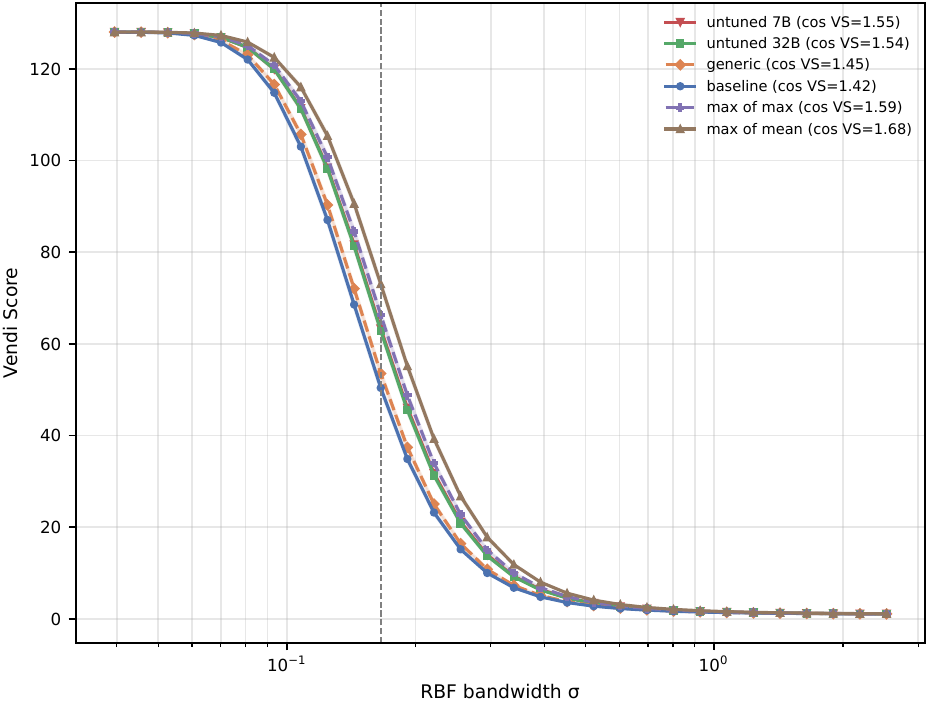}
  \caption{Average Vendi Score (effective number of distinct rollouts) of the generated chains of thought vs.\ the RBF kernel bandwidth $\sigma$, per method, averaged over the DeepMath val-hard held-out set. The methods are maximally separated at $\sigma=0.166$ (spread $=22.7$), the bandwidth used in \cref{tab:vendi_diversity}.}
  \label{fig:vendi_diversity}
\end{figure}

\textbf{Kernel bandwidth.}
The bandwidth $\sigma$ sets the scale at which two chains of thought count as similar, so we sweep it and plot the average VS of each method against $\sigma$ (\cref{fig:vendi_diversity}). We summarize each method at the bandwidth that maximizes the spread between methods (the difference between the largest and smallest average VS), $\sigma=0.166$, where the spread reaches $22.7$ points and the methods are best separated.

\textbf{Results.}
\Cref{tab:vendi_diversity} reports the average VS at this bandwidth, and diversity increases exactly in the order our exploration story predicts. Naive repeated sampling is the least diverse ($50.4$), and perturbing the prompt with generic mathematical text barely helps ($53.5$). Conditioning on sampled concepts raises diversity substantially, whether the concepts come from the untuned 7B or 32B generator ($63.3$ and $62.9$, essentially tied). Training the concept generator increases it further: max-of-max reaches $66.3$, and our best method, max-of-mean, is the most diverse at $73.0$---a $45\%$ increase in effective distinct rollouts over the naive baseline. This ordering mirrors the pass@k gains in \cref{tab:passk_ci}: methods that explore more diversely also solve more problems, supporting our claim that the gains come from broader exploration rather than prompt perturbation alone.

\begin{table}[!ht]
\centering
\caption{\label{tab:vendi_diversity}Average Vendi Score of the generated chains of thought (embedded with Qwen3-Embedding-8B, RBF kernel at bandwidth $\sigma=0.166$) on the DeepMath val-hard held-out set. Higher indicates more diverse reasoning. Best in \textbf{bold}, second best in \emph{italic}.}
\begin{tabular}{lc}
\toprule
Method & Vendi Score \\
\midrule
Naive baseline & $50.38$ \\
Generic prompt modification & $53.54$ \\
Untuned CG (7B) & $63.33$ \\
Untuned CG (32B) & $62.85$ \\
max-of-max & $\mathit{66.32}$ \\
max-of-mean & $\mathbf{73.04}$ \\
\bottomrule
\end{tabular}
\end{table}

\subsection{Effect of Longer Training}\label{app:more_steps}
\begin{figure}[t]
  \centering
  \includegraphics[width=0.4\linewidth]{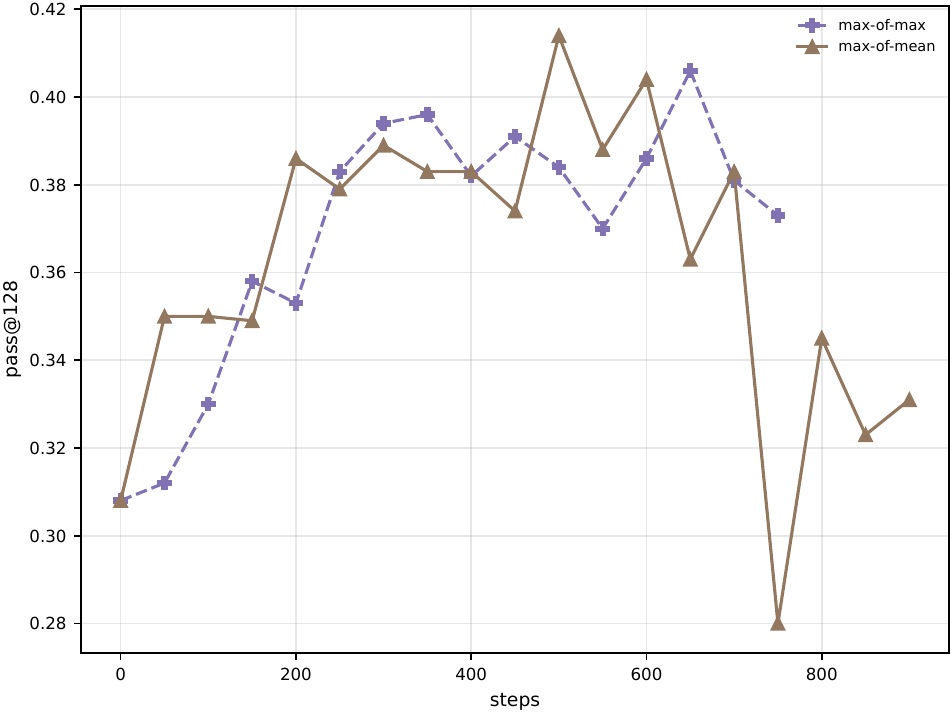}
  \caption{Held-out pass@128 on the DeepMath val-hard set over the course of training, for the two reward-aggregation objectives (max-of-max up to $750$ steps, max-of-mean up to $900$).}
  \label{fig:more_steps}
\end{figure}

Our main results (\cref{tab:main_results}) train the concept generator for $200$ steps. Because each step is dominated by the $G\times B$ answer-generator rollouts needed to score the sampled concept trajectories (\cref{sec:ablations}), we check whether this training length leaves performance on the table by continuing training well past it: max-of-max up to $750$ steps and max-of-mean up to $900$ (\cref{fig:more_steps}).

Performance largely saturates after the first few hundred steps. Both objectives climb steeply over the first $\sim\!200$ steps and then plateau in a noisy band around $0.38$--$0.40$ pass@128, with no sustained improvement thereafter. They differ mainly in how they reach this plateau: max-of-mean rises faster and is ahead at $200$ steps ($0.386$ vs.\ $0.353$ pass@128), whereas max-of-max catches up by $\sim\!300$ steps, after which the two are indistinguishable within the run-to-run noise. The slower early progress of max-of-max is expected: its trajectory reward is binary ($1$ if any concept in the trajectory yields a correct answer and $0$ otherwise) which, compounded by sampling noise, makes credit assignment harder and the learning signal less informative than the graded per trajectory success rate used by max-of-mean. Neither ever pulls clearly above the level already reached around $200$ steps.

The two objectives do differ in late-training stability. max-of-max stays within the plateau band through $750$ steps, whereas max-of-mean grows increasingly volatile and collapses sharply around step $750$ (pass@128 dropping to $\approx\!0.28$), recovering only partially afterwards. This collapse reflects a training instability rather than genuine overfitting: around the same point the policy entropy rises sharply and the concept generator increasingly emits over-long trajectories that hit the generation-length limit and are truncated before producing a stop token (the truncation rate climbs to $\approx\!60\%$ near step $720$), alongside a growing KL divergence from the reference policy. Such truncated concept trajectories are malformed and score poorly downstream, dragging pass@128 down. Together, these observations justify our $200$-step schedule: it captures essentially all of the attainable gain at a fraction of the compute, and stopping early avoids the instability that eventually affects max-of-mean.

\subsection{Qualitative Examples: Concepts Before and After Training}\label{app:concept_examples}
To illustrate qualitatively how training reshapes the generated concepts, we show three held-out problems for which concepts from the untuned 7B generator fail to elicit a correct AG answer, whereas concepts from the trained max-of-mean generator succeed (\cref{ex:concepts_ex1,ex:concepts_ex2,ex:concepts_ex3}). Ground-truth answers are shown only for reference and are not included in either the concept generator or answer generator prompts.

For each problem, we show three concepts from the untuned generator followed by three from the trained generator. In these examples, the trained concepts are more concrete and problem specific, providing a particular substitution, construction, or explicit line of attack, whereas the untuned concepts more often state generic strategies or name a relevant theorem without a usable plan.

\begin{examplebox}[ex:concepts_ex1]{Distributing balls with capacity constraints}
\textbf{Problem.} Find the generating function for the number of ways to distribute \( n \) identical juggling balls to five different jugglers, where each juggler receives at most seven balls. Using this generating function, determine the number of ways to distribute 25 balls.

\emph{Ground truth:} 926.

\medskip
\begin{untunedbox}
\begin{itemize}\setlength{\itemsep}{3pt}\setlength{\topsep}{0pt}
  \item Use the multinomial theorem to expand the generating function.
  \item Consider the stars and bars method with constraints to derive the generating function.
  \item Use the principle of inclusion-exclusion to count the valid distributions.
\end{itemize}
\end{untunedbox}

\begin{trainedbox}
\begin{itemize}\setlength{\itemsep}{3pt}\setlength{\topsep}{0pt}
  \item Use a generating function approach.
We can construct a generating function for the number of ways to distribute the balls under the given constraint. Each juggler can have 0, 1, 2, 3, 4, 5, 6, or 7 balls, so the generating function for one juggler is \(1 + x + x^2 + x^3 + x^4 + x^5 + x^6 + x^7\). Since there are five jugglers, the generating function for the distribution of all balls is the fifth power of this function.
  \item Explore a combinatorial approach.
Instead of using generating functions, consider a combinatorial argument. The problem can be translated into finding the number of non-negative integer solutions to the equation \(x_1 + x_2 + x_3 + x_4 + x_5 = 25\) with \(0 \leq x_i \leq 7\). This can be approached using the principle of inclusion-exclusion.
  \item Consider a recursive approach.
Define a recursive function \(f(n, k)\) that represents the number of ways to distribute \(n\) balls among \(k\) jugglers such that each juggler gets at most 7 balls. Use the recursive relation \(f(n, k) = \sum_{i=0}^{7} f(n-i, k-1)\) with base cases \(f(0, k) = 1\) and \(f(n, 0) = 0\) for \(n > 0\).
\end{itemize}
\end{trainedbox}
\end{examplebox}

\begin{examplebox}[ex:concepts_ex2]{Summing repeated integrals}
\textbf{Problem.} For non-negative integers $n$, define a function $h_n(x)$ as follows:

$$h_n\colon [0, 1]\to\mathbb{R}, \quad h_n(x)=\int_0^x h_{n-1}(t)dt\quad (n>0),\quad h_0(x) = x^2$$

Compute the sum $\displaystyle\sum_{n=0}^\infty h_n(x)$. Express your answer in terms of $x$.

\emph{Ground truth:} $2(e^x - x - 1)$.

\medskip
\begin{untunedbox}
\begin{itemize}\setlength{\itemsep}{3pt}\setlength{\topsep}{0pt}
  \item Recognize the pattern in integrals and derivatives.
  \item Use the exponential function series expansion.
  \item Apply induction to confirm the pattern of \( h_n(x) \).
\end{itemize}
\end{untunedbox}

\begin{trainedbox}
\begin{itemize}\setlength{\itemsep}{3pt}\setlength{\topsep}{0pt}
  \item Analyze the recursive relationship.
We can start by computing the first few terms of the sequence explicitly to identify a pattern. For $n=1$, we have:
\[ h_1(x) = \int_0^x h_0(t) \, dt = \int_0^x t^2 \, dt = \frac{x^3}{3}. \]

For $n=2$:
\[ h_2(x) = \int_0^x h_1(t) \, dt = \int_0^x \frac{t^3}{3} \, dt = \frac{x^4}{12}. \]

This suggests a general form for $h_n(x)$ might be of the form $\frac{x^{n+2}}{n! \cdot k_n}$ for some sequence $k_n$.
  \item Use generating functions.
Define a generating function $H(x, y) = \sum_{n=0}^\infty h_n(x) y^n$. Then, the recursive relation can be translated into a functional equation for $H(x, y)$. For example:
\[ H(x, y) = h_0(x) + y \int_0^x H(t, y) \, dt. \]
  \item Check for a closed-form solution using differentiation.
Differentiate the recursive relation with respect to $x$ and see if a pattern emerges. For example, if we differentiate $h_n(x) = \int_0^x h_{n-1}(t) \, dt$ with respect to $x$, we get:
\[ h_n'(x) = h_{n-1}(x). \]
This might help in finding a relation between $h_n(x)$ and $h_{n-1}(x)$.
\end{itemize}
\end{trainedbox}
\end{examplebox}

\begin{examplebox}[ex:concepts_ex3]{Constant-speed elevators (relative motion)}
\textbf{Problem.} Four elevators in a skyscraper, each a different color (red, blue, green, and yellow), move at constant speeds in various directions. The following events were observed: at $36$ seconds the red elevator caught up with the blue one; at $42$ seconds the red elevator passed the green one; at $48$ seconds the red elevator passed the yellow one; at $51$ seconds the yellow elevator passed the blue one; at $54$ seconds the yellow elevator caught up with the green one. Determine the time, in seconds from the start, when the green elevator will pass the blue one.

\emph{Ground truth:} $46$.

\medskip
\begin{untunedbox}
\begin{itemize}\setlength{\itemsep}{3pt}\setlength{\topsep}{0pt}
  \item Determine the relative speeds between pairs of elevators.
  \item Construct a system of equations based on the observed events.
  \item Plot the positions of the elevators as functions of time.
\end{itemize}
\end{untunedbox}

\begin{trainedbox}
\begin{itemize}\setlength{\itemsep}{3pt}\setlength{\topsep}{0pt}
  \item Introduce a coordinate system where one elevator is at rest to simplify the problem. Choose the blue elevator as the reference frame. Rewrite the positions and speeds of the other elevators relative to the blue elevator. This can simplify the equations since the blue elevator's position is constant in this frame.
  \item Solve for the initial positions and velocities of the elevators. Given that the red elevator catches up with the blue at $36$ seconds, we can use this event to write an equation for the initial positions and velocities. Similarly, use the other events to form a system of linear equations to solve for the unknowns.
  \item Consider the scenario where the green elevator passes the blue elevator. Set up an equation where the position of the green elevator equals the position of the blue elevator and solve for the time $t$.
\end{itemize}
\end{trainedbox}
\end{examplebox}

\end{document}